\documentclass[10pt,twocolumn,letterpaper]{article}

\usepackage{cvpr}              
\usepackage{adjustbox}
\usepackage{tabularx}
\usepackage{amssymb}   
\usepackage{pifont}    
\usepackage{multirow}
\usepackage{cuted}   
\usepackage{caption} 

\definecolor{cvprblue}{rgb}{0.21,0.49,0.74}
\usepackage[pagebackref,breaklinks,colorlinks,allcolors=cvprblue]{hyperref}
\usepackage{xcolor}
\usepackage{tcolorbox}
\usepackage{listings}
\usepackage{etoolbox}
\usepackage{amsmath, amssymb}

\usepackage{caption}   

\definecolor{best}{rgb}{1.0, 0.85, 0.6}
\definecolor{second}{rgb}{0.7, 0.9, 1.0}
\definecolor{sh_blue}{rgb}{0,0.60,0.93}
\definecolor{sh_gray2}{rgb}{1,0.89,0.75}
\definecolor{lyellow}{rgb}{1,0.63,0.098}
\definecolor{lred}{rgb}{0.906,0.42,0.32}
\definecolor{color3}{rgb}{0.95,0.95,0.95}
\definecolor{mygray}{gray}{.9}
\definecolor{genhaze}{rgb}{0.60, 0.57, 0.79}
\definecolor{bluegreen}{rgb}{0.44, 0.64, 0.77}
\definecolor{gray_venue}{rgb}{0.53,0.52,0.52}
\definecolor{color5}{rgb}{1,0.96,0.88}

\lstdefinelanguage{json}{
  basicstyle=\ttfamily\small,
  breaklines=true,
  keepspaces=true,
  showstringspaces=false,
  morecomment=[l]{//},
  morestring=[b]",
  literate=
    *{0}{{{\color{black}{0}}}}1
     {1}{{{\color{black}{1}}}}1
     {:}{{{\color{black}{:}}}}1
     {,}{{{\color{black}{,}}}}1
     {\{}{{{\color{black}{\{}}}}1
     {\}}{{{\color{black}{\}}}}}1
     {[}{{{\color{black}{[}}}}1
     {]}{{{\color{black}{]}}}}1
}

\tcbuselibrary{listingsutf8,skins}
\tcbset{
    examplebox/.style={
        enhanced,
        colback=sh_blue!5,
        colframe=sh_blue,
        coltitle=sh_blue,
        fonttitle=\bfseries,
        boxrule=0.6mm,
        sharp corners,
        attach boxed title to top left={
            yshift=-2mm,
            xshift=5mm
        },
        boxed title style={
            colframe=sh_blue,
            colback=white,
            sharp corners,
            boxrule=0.6mm
        },
        before upper={\small\rmfamily\color{black}},
        fontupper=\small,
        boxsep=6pt,
        left=10pt,
        right=10pt,
        top=8pt,
        bottom=8pt
    }
}

\newtcolorbox[auto counter, number within=section]{example}[2][]{%
    examplebox,
    title=Prompt~\thetcbcounter~(#2),
    #1
}

\def\paperID{***} 
\def\confName{CVPR}
\def\confYear{2020}

\begin{document}

\title{UltraFlux: Data-Model Co-Design for High-quality Native 4K Text-to-Image Generation across Diverse Aspect Ratios}

\author{
\begin{tabular}{ccc}
Tian Ye\thanks{Equal contribution, \emph{UltraFlux} project leader: Tian Ye} &
Song Fei\footnotemark[1] &
Lei Zhu\thanks{Corresponding author.} \\
HKUST(GZ) & HKUST(GZ) & HKUST, HKUST(GZ) \\
{\tt\small tye610@connect.hkust-gz.edu.cn} &
{\tt\small sfei285@connect.hkust-gz.edu.cn} &
{\tt\small leizhu@ust.hk}
\end{tabular}
\\[1em]
\small \textbf{Project}: \url{https://w2genai-lab.github.io/UltraFlux/} \\
\small \textbf{Code}: \url{https://github.com/W2GenAI-Lab/UltraFlux}
\vspace*{-3.0cm}
}

\maketitle
\vspace{-0.7cm}
\begin{strip}
\vspace*{-0.8em} 
\centering
\includegraphics[width=\textwidth]{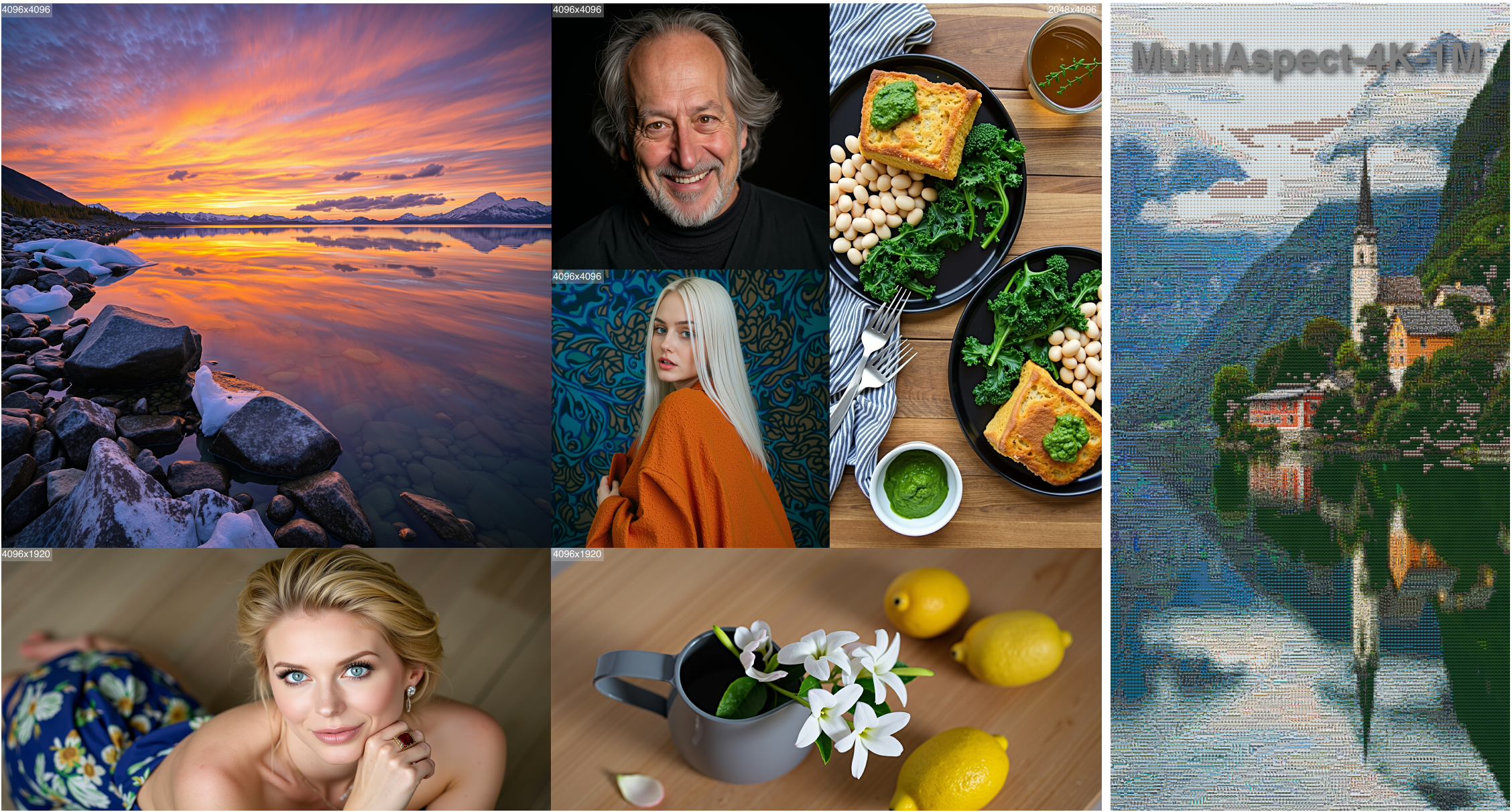}
\vspace*{-0.8em} 
\captionsetup{type=figure,font=small} 
\captionof{figure}{\textbf{Left:} UltraFlux generates photorealistic 4K images across diverse aspect ratios and topics while maintaining high aesthetic quality and faithful content depiction with a single unified text-to-image model. \textbf{Right:} Our MultiAspect-4K-1M is a large-scale high-quality dataset for 4K image synthesis.}
\label{fig:teaser}
\vspace*{-0.8em} 
\end{strip}

\begin{abstract}
Diffusion transformers have recently delivered strong text-to-image generation around 1K resolution, but we show that extending them to \emph{native 4K} across diverse aspect ratios exposes a tightly coupled failure mode spanning positional encoding, VAE compression, and optimization. Tackling any of these factors in isolation leaves substantial quality on the table. We therefore take a data--model co-design view and introduce \emph{UltraFlux}, a Flux-based DiT trained natively at 4K on \emph{MultiAspect-4K-1M}, a 1M-image 4K corpus with controlled multi-AR coverage, bilingual captions, and rich VLM/IQA metadata for resolution- and AR-aware sampling. On the model side, UltraFlux couples (i) \emph{Resonance 2D RoPE with YaRN} for training-window-, frequency-, and AR-aware positional encoding at 4K; (ii) a simple, non-adversarial VAE post-training scheme that improves 4K reconstruction fidelity; (iii) an \emph{SNR-Aware Huber Wavelet} objective that rebalances gradients across timesteps and frequency bands; and (iv) a \emph{Stage-wise Aesthetic Curriculum Learning} strategy that concentrates high-aesthetic supervision on high-noise steps governed by the model prior. Together, these components yield a stable, detail-preserving 4K DiT that generalizes across wide, square, and tall ARs. On the Aesthetic-Eval@4096 benchmark and multi-AR 4K settings, UltraFlux consistently outperforms strong open-source baselines across fidelity, aesthetic, and alignment metrics, and—with a LLM prompt refiner—matches or surpasses the proprietary Seedream~4.0. 
\end{abstract}

\section{Introduction}
\label{sec:intro}
Diffusion transformers (DiTs)~\cite{peebles2023scalable,batifol2025flux,esser2024scaling,chen2024pixart,xie2024sana} have recently pushed text-to-image generation to impressive quality around 1K resolution, enabled by efficient backbones, token compression, and carefully tuned training pipelines \cite{chen2024pixart,xie2024sana}. However, extending these systems to \emph{native 4K} while supporting a broad spectrum of aspect ratios (ARs) is not a simple matter of scaling resolution. At 4096$\times$4096 and beyond, we empirically observe three coupled challenges: (i) \emph{positional representation and AR extrapolation}, where 2D rotary embeddings calibrated on a single training window can drift or alias under large changes in resolution and AR \cite{peng2023yarn,zhang2024hirope}; (ii) \emph{high-frequency fidelity under VAE compression}, where higher downsampling factors improve throughput but tend to erase fine structures that dominate 4K perception \cite{xie2024sana,zhang2025diffusion}; and (iii) \emph{4K-aware optimization}, where gradient contributions become heavily skewed across timesteps and frequency bands, making standard objectives poorly matched to the statistics of 4K latents \cite{hang2023efficient,zhang2025diffusion}. These factors interact: the choice of positional scheme, VAE compression ratio, and training objective jointly determines whether a model can remain stable and detailed across native 4K resolutions and diverse ARs.

On the model side, several scaling strategies partially address these issues but leave the overall design space fragmented. Training-free high-resolution methods mitigate tiling artifacts and duplication at inference time, yet largely preserve the underlying positional encoding and were not designed for systematic multi-AR extrapolation \cite{zhang2024hidiffusion,huang2024fouriscale}. Decoder-side approaches based on global–local fusion or tiled diffusion improve size flexibility but introduce new failure modes, such as coherence gaps across tiles or heavy reliance on a global prior for consistency \cite{haji2024elasticdiffusion,bar2023multidiffusion}. Native-4K systems~\cite{yu2025ultra,liu2024clear} demonstrate that carefully engineered backbones can make 4K training tractable \cite{chen2024pixart,xie2024sana}, but most emphasize token/architecture efficiency and treat \emph{positional robustness, VAE compression, and loss design} as largely orthogonal choices rather than a jointly optimized 4K regime.

Progress at 4K is further constrained by the data itself. Public 4K corpora are typically modest in scale (on the order of $10^4$–$10^5$ images), heavily biased toward near-square ARs and landscape-centric content, and curated with early CLIP-based aesthetic predictors. For example, Aesthetic-4K takes an important step by assembling high-quality 4K image–text pairs with GPT-4O captions \cite{zhang2025diffusion}, yet its scale and AR coverage remain limited for studying \emph{resolution–AR coupling}, and its subject distribution under-represents human-centric scenes. More critically, existing 4K datasets rarely provide the \emph{structured metadata} needed for modern 4K training. As a result, practitioners have limited control over sampling data slices tailored to specific training regimes (e.g., high-detail or high-aesthetic subsets), and it becomes difficult to perform fine-grained aesthetic or AR-conditioned analyses.

On the optimization and adaptation side, recent work explores complementary—but still incomplete—directions. Wavelet-aware objectives at native resolution improve fidelity on strong backbones by better emphasizing high-frequency content \cite{zhang2025diffusion}, yet they typically combine simple quadratic or perceptual penalties and thus remain vulnerable to cross-scale dominance of low-frequency energy. Latent-space super-resolution and self-cascade schemes sharpen details beyond the original training resolution and reduce the cost of high-resolution transfer \cite{jeong2025latent,guo2024make}, but they operate as post-hoc adapters on fixed backbones and do not resolve the underlying trade-off between VAE compression and 4K reconstruction fidelity. In parallel, timestep curricula adjust noise sampling while holding the data distribution fixed, and aesthetic post-training applies high-aesthetic data uniformly across timesteps, leaving unexplored the regime where \emph{high-noise steps—those most governed by the model prior—are selectively sculpted by high-aesthetic supervision}. Finally, existing RoPE interpolation and NTK-style scaling strategies are primarily developed for 1D sequence length extrapolation, and provide little guidance for \emph{2D token grids at native 4K under strongly varying ARs}, where misaligned phase behavior manifests as ghosting, drift, and striping artifacts. Altogether, native 4K multi-AR generation still lacks a unified framework that couples: (i) a large-scale, multi-AR, content-diverse, VLM-curated 4K corpus with rich metadata; (ii) an efficient, non-adversarial VAE post-training strategy that improves 4K reconstruction without sacrificing throughput; (iii) an SNR-Aware Huber Wavelet Training Objective and a stage-wise aesthetic curriculum matched to 4K statistics; and (iv) a training-window aware, band-aware, and AR-aware positional encoding scheme. In this work, we explicitly target this data–model co-design space. 

Concretely, we make the following contributions:
\begin{itemize}
    \item \textbf{MultiAspect-4K-1M: a large-scale, multi-AR, aesthetically curated 4K corpus.} We construct a 1M-scale 4K dataset with native 4K and near-4K resolution, controlled aspect-ratio coverage, and a dual-channel pipeline that debiases landscape-heavy sources toward human-centric content. Each image is accompanied by decoupled VLM-based quality and aesthetic scores, classical IQA signals, bilingual captions, and subject tags, providing the structured metadata needed for data--model co-design.
    
    \item \textbf{UltraFlux: a data--model co-designed DiT for native 4K multi-AR generation.} We train a Flux-based backbone on MultiAspect-4K-1M with a co-designed recipe that couples (i) \emph{Resonance 2D RoPE with YaRN} for training-window aware, band-aware, and AR-aware positional encoding, (ii) an \emph{SNR-Aware Huber Wavelet Training Objective} tailored to 4K latents, (iii) a \emph{Stage-wise Aesthetic Curriculum Learning (SACL)} scheme that concentrates high-aesthetic supervision on high-noise steps, and (iv) a simple, non-adversarial, data-efficient VAE post-training procedure that improves Flux VAE reconstructions at 4K. Together, these components yield a stable, detail-preserving DiT for native 4K synthesis across diverse ARs.
    
    \item \textbf{State-of-the-art native 4K performance.} On standard 4K benchmarks and popular metrics covering fidelity, aesthetic quality, and text alignment, UltraFlux consistently outperforms strong 4K baselines, including recent native-4K and training-free scaling methods.
\end{itemize}

\section{Related Work}
This section reviews approaches to scaling text-to-image diffusion models to high-resolution T2I, native‑4K and diverse aspect ratios. We group prior work into three lines: training‑free inference‑time scaling, lightweight adaptations (e.g., latent super‑resolution and self‑cascade), and native‑4K training with 4K‑capable backbones.

\noindent\textbf{Training-Free High-Resolution Scaling.} Training-free strategies extend pre-trained 512--1K models to 2K/4K and diverse aspect ratios by modifying inference-time computation, without re-training. \emph{HiDiffusion} diagnoses duplication and quadratic self-attention costs at high resolutions and introduces a resolution-aware U-Net and windowed attention to improve quality and speed \cite{zhang2024hidiffusion}. \emph{FouriScale} approaches ultra-high resolution from the frequency view via Fourier-domain low-pass guidance and dilated convolutions, improving global structure while injecting high frequencies \cite{huang2024fouriscale}. 
These approaches are effective for quick scaling, yet commonly \emph{keep the original positional scheme unchanged}, which leaves \emph{positional extrapolation stability under extreme ARs} only partially addressed \cite{zhang2024hidiffusion,huang2024fouriscale,haji2024elasticdiffusion,bar2023multidiffusion}.

\noindent\textbf{Lightweight Adaptation: Latent SR and Self-Cascade Models.}
Lightweight adaptations improve high-resolution quality with minimal cost by augmenting the sampling pipeline or attaching small modules. \emph{LSRNA} maps low-res latents to a high-res manifold via latent-space super-resolution and injects region-wise noise to restore high-frequency detail without retraining the base model \cite{jeong2025latent}. \emph{Self-Cascade Diffusion} integrates low-resolution generation into the high-resolution denoising process and optionally fine-tunes small multi-scale upsamplers, achieving rapid 4K adaptation at a fraction of full fine-tuning cost \cite{guo2024make}. While these methods markedly sharpen details and reduce adaptation overhead, they typically \emph{inherit the original positional scheme}, leaving \emph{AR-generalized extrapolation} under-explored \cite{jeong2025latent,guo2024make}.

\noindent\textbf{Native 4K Training and 4K-Capable Foundation Models.}
A complementary direction trains or fine-tunes models directly at native‑4K and curates high-quality 4K corpora. \emph{Diffusion-4K} introduces Aesthetic-4K and a wavelet-based fine-tuning scheme that improves fidelity and prompt alignment on large modern backbones \cite{zhang2025diffusion}. Meanwhile, efficient backbones such as \emph{PixArt-$\Sigma$} (token-compression attention) and \emph{Sana} (32$\times$ VAE with linear-attention DiT) make 4096$\times$4096 synthesis computationally feasible at small model scales \cite{chen2024pixart,xie2024sana}. Despite these advances, public corpora remain limited in \emph{scale and AR diversity}, constraining systematic study of \emph{resolution--AR coupling}.Moreover, end‑to‑end methodologies for stable native‑4K training are under‑documented and fragmented across implementations, slowing progress and curbing real‑world adoption while masking the gains of true 4K training.
A practical distinction concerns \emph{native} versus \emph{upscaler-based} 4K. Unlike platform services that reach 4K primarily via 2$\times$/4$\times$ upscalers (e.g., Midjourney~\cite{MidJourney}; Google Imagen~\cite{Imagen} exposes a dedicated upscaler; Ideogram offers a 2$\times$ Upscale endpoint~\cite{ideogram2025upscale}), recent closed-source leaders such as \emph{Seedream~4}~\cite{seedream2025seedream} explicitly support \emph{multi-AR, native‑4K} generation within a unified T2I/editing architecture. This distinction matters: cascade upscaling pipelines couple low-resolution synthesis with a separate restoration prior, conflating high-frequency fidelity and positional extrapolation, whereas native‑4K training compels the backbone to learn long-range dependencies and cross-AR spatial alignment directly. \textit{Therefore, we treat native‑4K as a distinct training/evaluation regime and design both our data (MultiAspect-4K-1M) and recipe (UltraFlux) to isolate the gains of true 4K training from post-hoc upscaling.}


\section{Method: Data–Model Co-Design for Native 4K Multi-AR Generation}
\subsection{MultiAspect-4K-1M Dataset}
\label{sec:dataset}
\begin{figure}[t]
\centering
\vspace{-0.6cm}
\includegraphics[page=1, width=0.9\linewidth]{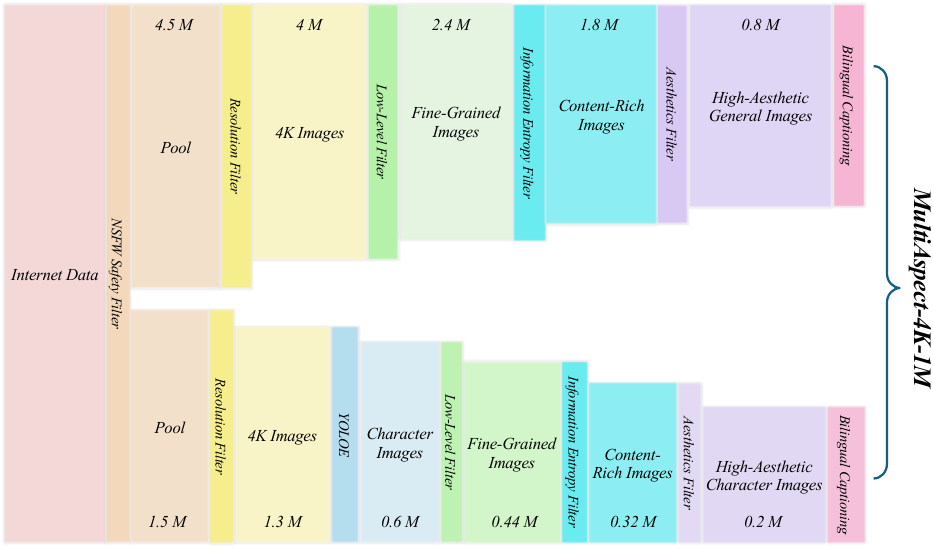}
\vspace{-0.3cm}
\caption{Data Pipeline overview.}
\label{fig:dataset_overview}
\vspace{-0.7cm}
\end{figure}

\noindent\textbf{Design goals and scope.}
Public 4K corpora for text-to-image training remain modest in scale (typically below \(10^5\) images) and are usually curated with early CLIP-based aesthetic predictors such as LAION-Aesthetic. While these datasets already achieve reasonably good visual quality, their \emph{aspect-ratio (AR) coverage} is coarse and imbalanced---only a few popular ARs are well-populated at native 4K---and the textual side (captions and aesthetic/quality supervision) is constrained by legacy CLIP-only scoring. Our data design therefore targets three complementary gaps: (i) \emph{broad multi-AR coverage} at native 4K to avoid overfitting to a small set of AR buckets; (ii) \emph{refreshed supervision quality}, coupling modern VLM-based quality/aesthetics estimators  instead of relying solely on legacy CLIP-based predictors; and (iii) \emph{distribution debiasing} that compensates for the over-representation of landscapes and the under-representation of human-centric content in existing 4K sources. We adopt a \emph{VLM-driven} filtering strategy---semantic quality via \emph{Q-Align}~\cite{wu2023qalign} and aesthetics via \emph{ArtiMuse}~\cite{cao2025artimusefinegrainedimageaesthetics}---complemented by interpretable classical signals (flatness and information entropy), and a dedicated \emph{character augmentation} branch to improve recall for human subjects. Figure~\ref{fig:dataset_overview} sketches the two-channel pipeline and the final merge.

\noindent\textbf{Sources and overall structure.}
After an NSFW safety check, we curate from a pool of approximately \(6\)M high-resolution images whose subject distribution is skewed toward landscapes. To operationalize the goals in Sec.~\ref{sec:dataset}, we adopt a \emph{dual-channel} pipeline: (i) a \textbf{general, AR-aware curation} path that enforces native/near-4K resolution and broad aspect-ratio (AR) coverage while filtering for quality and aesthetics; and (ii) a \textbf{human-centric augmentation} path that restores the underrepresented \emph{character} category via open-vocabulary detection. The two channels are merged after de-duplication and with consistent metadata (resolution/AR, VLM scores, classical signals, caption, subject tags), yielding \emph{1M} images. Figure~\ref{fig:dataset_overview} provides a high-level overview and stage-wise retention.

\begin{figure}[t]
\centering
\vspace{-0.3cm}
\includegraphics[page=1, width=0.75\linewidth]{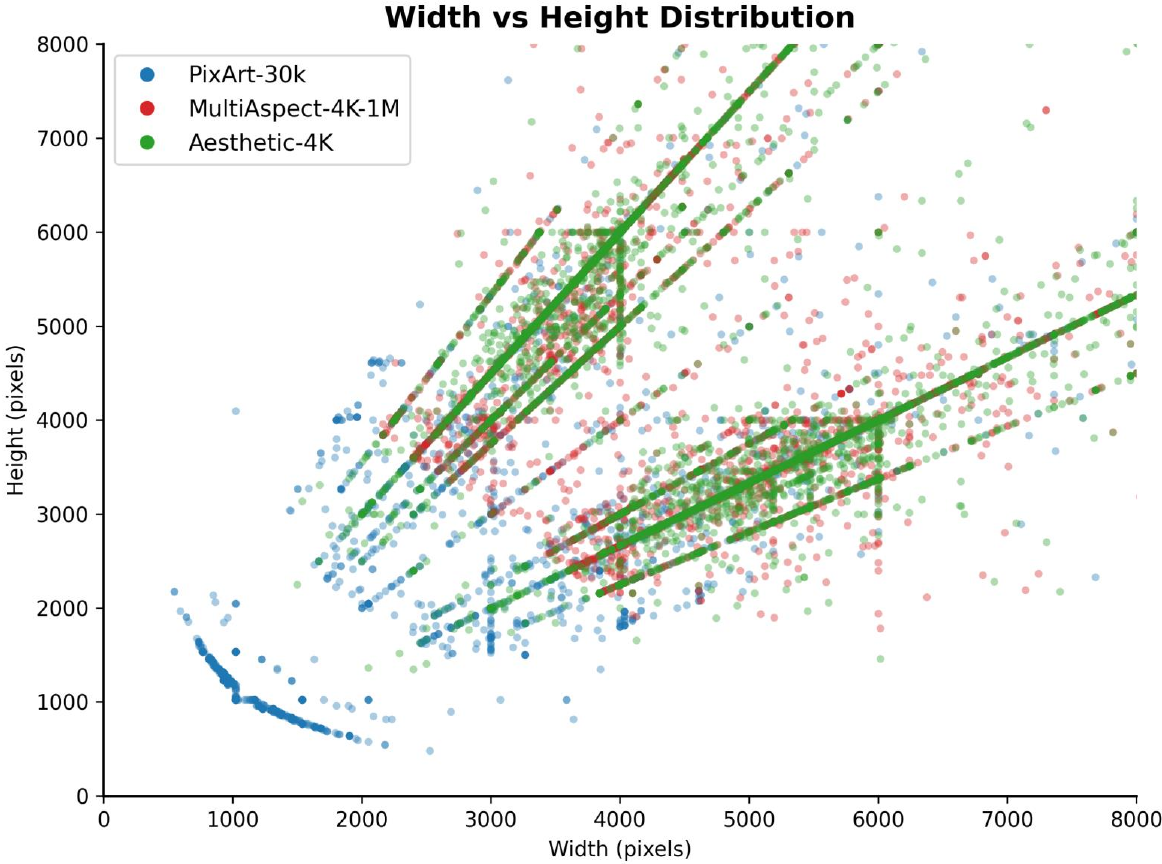}
\vspace{-0.3cm}
\captionsetup{font=scriptsize} 
\caption{Dataset aspect and resolution analysis. All datasets use 10k samples. MultiAspect-4K-1M has a broader aspect ratio distribution.}
\label{fig:dataset_aspect}
\vspace{-0.5cm}
\end{figure}

\noindent\textbf{VLM-based Filtering for HQ 4K Data.}
We begin with a safety screen, then enforce a \emph{pixel-count threshold} as resolution filtering stage—images must have at least 3840 $\times$ 2160 total pixels; we \emph{preserve each image’s native aspect ratio without any resizing}. This keeps the corpus artifact-free while naturally retaining a wide spectrum of ARs (e.g., 1:1, 16:9, 3:2, 4:3, 9:16), enabling transparent auditing of AR coverage. The resulting resolution and aspect-ratio distribution across 4K corpora is visualized in Figure~\ref{fig:dataset_aspect}. On this scaffold, we \emph{decouple} \textbf{quality} from \textbf{aesthetics}: for quality we adopt \emph{Q-Align}, a large multimodal model (LMM)-based visual scorer shown to deliver robust IQA judgments via discrete text-level supervision, and for aesthetics we use \emph{ArtiMuse}, a recent MLLM-based image aesthetics evaluator that provides numeric scores together with \emph{reasoned, expert-style explanations} (rather than score-only outputs). Classical, interpretable signals—\emph{flatness} and \emph{information entropy}—act as guardrails to suppress low-texture or overly smooth images that VLMs may tolerate, yielding a cleaner, high-frequency–preserving pool without sacrificing semantic clarity.

\begin{figure}[t]
\centering
\vspace{-0.6cm}
\includegraphics[page=2, width=0.9\linewidth]{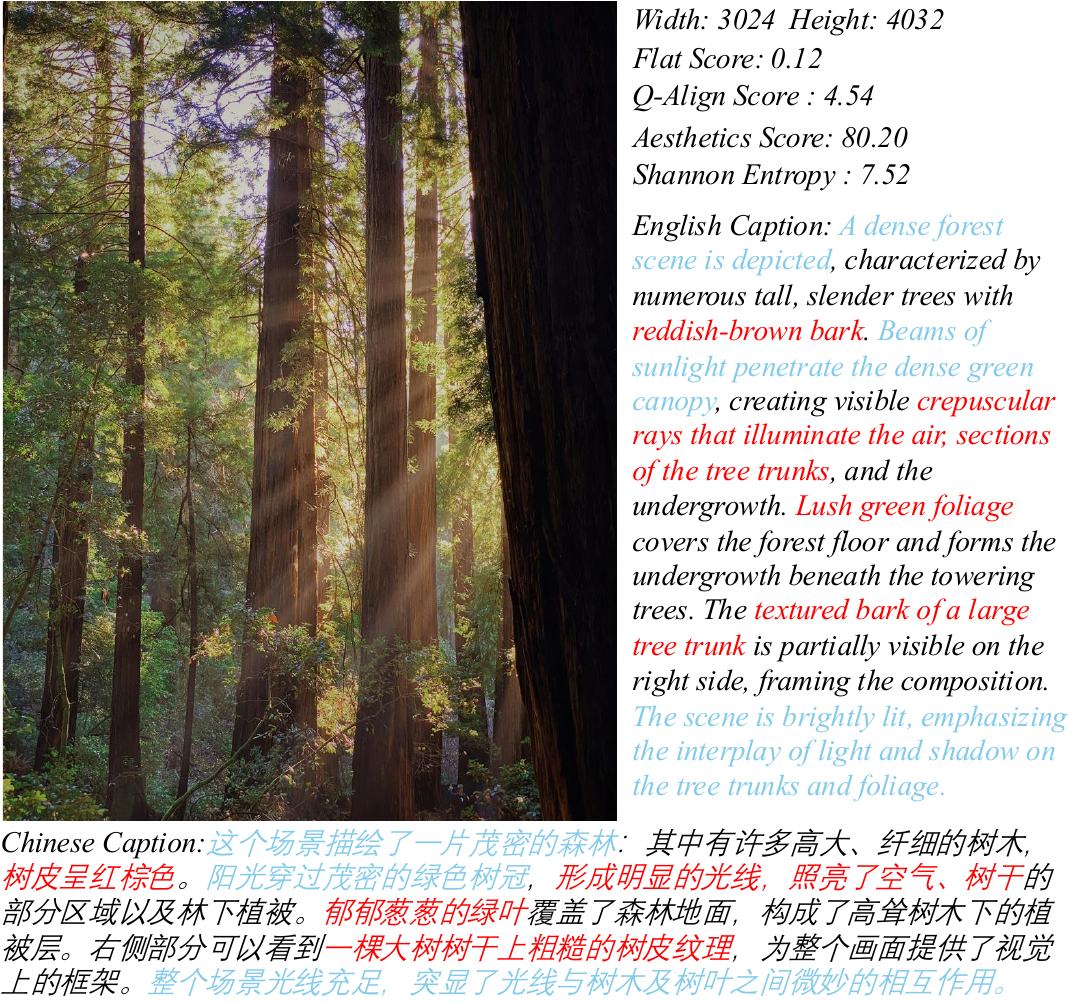}
\vspace{-0.3cm}
\captionsetup{font=scriptsize} 
\caption{Dataset example.}
\label{fig:dataset_example}
\vspace{-0.1cm}
\end{figure}

\noindent\textbf{Human-centric augmentation via open-vocabulary detector.}
To correct the chronic under-representation of people in 4K sources, we run a targeted augmentation path. Candidate images are collected via person-related retrieval under the same safety and resolution/AR checks. We then apply the same \emph{Q-Align} and \emph{ArtiMuse} filters, strengthened by \emph{information entropy} to suppress low-texture portraits. Crucially, we require \emph{structured evidence of human presence} using \textbf{YOLOE}~\cite{wang2025yoloerealtimeseeing}, a promptable open-vocabulary detector, which improves both recall and precision beyond fixed-class detectors. The accepted subset is merged into the main pool with a \texttt{character} flag.

Key statistics and comparisons to existing 4K and popular T2I datasets are summarized in Table~\ref{tab:dataset_stats}.

\begin{table}[htbp]
\centering
\vspace{-0.3cm}
\caption{Dataset statistical comparisons.}
\vspace{-0.3cm}
\label{tab:dataset_stats}
\resizebox{0.98\linewidth}{!}{%
\begin{tabular}{lccccc}
\toprule
Dataset & Number & Avg. Resolution & Avg. ArtiMuse & Avg. Caption Length & Bilingual Caption \\
\midrule
PixArt-30k~\cite{chen2024pixart} 
    & 30,000     & 2,531$\times$2,656 & 63.67 & 87.9 tokens & \ding{55} \\
Aesthetic-4K~\cite{zhang2025diffusion}     
    & 12,009     & 4,576$\times$4,837 & 63.49 & 31.0 tokens & \ding{55} \\
MultiAspect-4K-1M                           
    & 1,007,230  & 4,521$\times$4,703 & 64.59 & 125.1 tokens & \checkmark \\
\bottomrule
\end{tabular}
}
\vspace{-0.4cm}
\end{table}

\noindent\textbf{Bilingual captioning.}
Captioning is performed \emph{last}. For the retained set, we generate detailed captions with \emph{Gemini-2.5-Flash}, a production multimodal model suitable for fast, high-quality captioning; we then translate each caption into accurate Chinese with \emph{Hunyuan-MT-7B}~\cite{zheng2025hunyuan} to serve bilingual users. An example image with its metadata and bilingual captions is shown in Figure~\ref{fig:dataset_example}. 

The final \textit{MultiAspect-4K-1M} corpus comprises \emph{1M} 4K images with balanced coverage over standard AR buckets and a diversified subject mix (landscapes, people, objects). Each image includes resolution, \emph{Q-Align}, \emph{ArtiMuse}, flatness/entropy, English/Chinese caption, and the \texttt{character} tag. These fields are designed for transparent auditing and flexible data-model co-design: \emph{they can act as analysis tags and stratified sampling keys for text-to-image training.}

\subsection{UltraFlux: Scaling Flux to Native 4K Image Generation}
\label{sec:ultraflux}
With the data foundation in place, we now turn to the model side and build \emph{UltraFlux}. Rather than redesigning the DiT architecture, we keep the core Flux transformer intact and focus on three components that bottleneck 4K performance: the VAE, the positional representation, and the training objective and strategy. We first post-train an $F16$ VAE to recover fine details without giving up the efficiency gains of stronger compression, then introduce a Resonance 2D RoPE with a YaRN-style extrapolation scheme to stabilize attention across resolutions and aspect ratios. Finally, we couple an SNR-aware Huber wavelet loss with a stage-wise aesthetic curriculum that concentrates learning on high-frequency structure and high-aesthetic examples. Together, these lightweight but targeted changes upgrade Flux into an efficient, high-fidelity 4K generator that can fully exploit \emph{MultiAspect-4K-1M}.

\subsubsection{VAE Post-training for High-resolution Reconstruction Fidelity}

A strong but efficient VAE is essential for practical native 4K image generation. The Flux backbone uses an $F8$ VAE (height/width downsampling by $8$), which at 4K yields a very large latent grid and makes sampling prohibitively slow: in our profiling, a single 4K image with $50$ diffusion steps takes on the order of $30$ minutes. Following Diffusion-4K~\cite{zhang2025diffusion}, we instead adopt an $F16$ VAE, halving the latent resolution while keeping comparable channel capacity, and focus on \emph{post-training} the decoder to improve high-resolution reconstruction fidelity.

We fine-tune the Flux $F16$ decoder on our curated \emph{MultiAspect-4K-1M} corpus to enhance fine-scale 4K detail. Ablations on loss design and training recipes lead to three key findings: \emph{(i)} explicitly targeting high-frequency content is essential—combining a wavelet reconstruction loss applied to high-frequency sub-bands with a feature-space perceptual loss consistently outperforms purely pixel-wise and perceptual objectives in sharpness and structural fidelity; \emph{(ii)} once reconstruction reaches a reasonable regime, an adversarial discriminator offers negligible benefit—the GAN loss saturates quickly, induces optimization instability, and fails to improve perceptual quality, so our final recipe omits the adversarial term and retains only wavelet, perceptual and $L_2$ losses; and \emph{(iii)} stringent data curation substantially reduces post-training cost—by applying a flatness filter to select a high-detail subset of \emph{MultiAspect-4K-1M}, we obtain the bulk of reconstruction gains within around $4\text{k}$ update steps, and observe that a few hundred thousand carefully screened, detail-rich images suffice to markedly upgrade the Flux $F16$ VAE without multi-day GAN training or tens of millions of samples. This lightweight post-training stage produces a decoder that preserves fine 4K structures while maintaining the throughput advantages of $F16$ compression, thereby enabling native 4K synthesis with both high fidelity and practical efficiency.

\subsubsection{Resonance 2D RoPE for Multi-AR 4K Extrapolation}
\label{sec:resonance-rope}
The official Flux backbone employs a fixed per-axis rotary spectrum with an optional \emph{global} NTK factor \cite{batifol2025flux}, following the standard RoPE formulation \cite{su2024roformer} but without band-specific treatment or training-window awareness.
Consequently, the frequencies do not adapt to the inference size $H{\times}W$, and phase grows purely with position, which empirically destabilizes multi–aspect-ratio generation at native 2K/4K.

\noindent\textbf{Flux baseline: 2D RoPE.}
Following Flux, we assign rotary embeddings independently along height and width.
For each axis $a\in\{H,W\}$ with channel size $d_a$ and $m_a=d_a/2$ complex pairs, we use rotary base $b>1$ and optional NTK factor $\eta\ge1$ to define per-axis frequencies
\begin{equation}
\omega^{(a)}_{k} \;=\; (b\cdot\eta)^{-\alpha^{(a)}_k},
\qquad 
\alpha^{(a)}_k \;=\; \frac{2k}{d_a},
\qquad 
k=0,\ldots,m_a-1,
\label{eq:flux-freq}
\end{equation}
and phases for a position $p=(p_H,p_W)$ (in patches)
\begin{equation}
\phi^{(a)}_{k}(p_a) \;=\; p_a\,\omega^{(a)}_{k}.
\label{eq:flux-phase}
\end{equation}
Writing each pair as $z^{(a)}_{k}=x^{(a)}_{2k-1}+i\,x^{(a)}_{2k}$, we apply the usual complex rotation
$\tilde{z}^{(a)}_{k} = z^{(a)}_{k}\,e^{\,i\,\phi^{(a)}_{k}(p_a)}$,
with wavelength (in patches) $\lambda^{(a)}_{k} = 2\pi / \omega^{(a)}_{k}$.

\noindent\textbf{Resonance 2D RoPE.}
Motivated by the Resonance RoPE idea for train-short-test-long generalization in LLMs \cite{wang2024resonance}, we reinterpret the 2D rotary spectrum on a \emph{finite} training window.
Let $L_H,L_W$ be the training-window lengths (in patches) along height and width, and let $\omega^{(a)}_{k}$ be the per-axis frequencies from Eq.~\eqref{eq:flux-freq}.
We define the number of cycles completed by component $(a,k)$ inside the training window as
\begin{equation}
r^{(a)}_{k} \;=\; \frac{L_a}{\lambda^{(a)}_{k}}
\;=\; \frac{L_a\,\omega^{(a)}_{k}}{2\pi}.
\label{eq:reson-cycles}
\end{equation}
We then \emph{snap} $r^{(a)}_{k}$ to the nearest nonzero integer:
\begin{equation}
\hat{r}^{(a)}_{k} \;=\; \max\!\Big(1,\ \big\lfloor r^{(a)}_{k}+\tfrac{1}{2}\big\rfloor\Big),
\label{eq:reson-integer}
\end{equation}
and replace $\omega^{(a)}_{k}$ by its integer-cycle projection
\begin{equation}
\hat{\omega}^{(a)}_{k} \;=\; \frac{2\pi\,\hat{r}^{(a)}_{k}}{L_a}.
\label{eq:reson-omega}
\end{equation}
The phase becomes
\begin{equation}
\hat{\phi}^{(a)}_{k}(p_a) \;=\; p_a\,\hat{\omega}^{(a)}_{k},
\label{eq:reson-phase}
\end{equation}
and we reuse the same complex rotation as in the Flux baseline, now driven by $\hat{\phi}^{(a)}_{k}$.
The snapped wavelength is $\hat{\lambda}^{(a)}_{k}=2\pi/\hat{\omega}^{(a)}_{k}=L_a/\hat{r}^{(a)}_{k}$.

\noindent\textbf{Discussion.}
Snapping $r^{(a)}*{k}$ to the nearest nonzero integer turns each rotary band into a finite-window ``standing wave'' on $[0,L_a]$ that completes an exact integer number of cycles and has matching phase at $p_a=0$ and $p_a=L_a$. In the original Flux spectrum, many bands traverse the training window with a fractional number of cycles, so reusing the same frequencies at larger resolutions or different ARs accumulates half-cycle phase error, which appears as spatial drift and faint striping, especially in high-frequency channels. By replacing $\omega^{(a)}*{k}$ in Eq.~\eqref{eq:flux-freq} with their resonant counterparts $\hat{\omega}^{(a)}_{k}$, Resonance 2D RoPE makes the spectrum explicitly training-window aware and prevents this fractional-cycle build-up, empirically reducing ghosting and striping artifacts when extrapolating to native-4K multi-AR grids.

\noindent\textbf{Resonance 2D RoPE with YaRN.}
Inspired by the YaRN scheme for length extrapolation of 1D RoPE \cite{peng2023yarn}, we further make the extrapolation \emph{band-aware}.
Let $a\!\in\!\{H,W\}$ index spatial axes.
Given the training-window length $L_a$ (in patches) and the \emph{resonant} frequency $\hat{\omega}^{(a)}_{k}$ with integer cycles $\hat{r}^{(a)}_{k}$ from the previous section, define the inference length $L'_a$ and the extrapolation scale $s_a=L'_a/L_a\!\ge\!1$.
We use a linear ramp to map each band:
\begin{equation}
\gamma(r;\alpha,\beta)=
\begin{cases}
0, & r<\alpha,\\[2pt]
\dfrac{r-\alpha}{\beta-\alpha}, & \alpha\le r\le \beta,\\[6pt]
1, & r>\beta,
\end{cases}
\qquad 0\le \alpha<\beta,
\label{eq:reson-yarn-ramp}
\end{equation}
and interpolate between position-interpolation scaling and no scaling using the resonant cycle count:
\begin{equation}
\omega^{(a)}_{k,\mathrm{yarn}}
\;=\;
\bigl(1-\gamma(\hat{r}^{(a)}_{k};\alpha,\beta)\bigr)\,\frac{\hat{\omega}^{(a)}_{k}}{s_a}
\;+\;
\gamma(\hat{r}^{(a)}_{k};\alpha,\beta)\,\hat{\omega}^{(a)}_{k}.
\label{eq:reson-yarn-omega}
\end{equation}
The phase is $\phi^{(a)}_{k,\mathrm{yarn}}(p_a)=p_a\,\omega^{(a)}_{k,\mathrm{yarn}}$, and the complex rotation is identical to the Flux baseline.
Compared to Flux's fixed spectrum with a single global NTK factor, Resonance 2D RoPE with YaRN first snaps frequencies to finite-window resonant modes and then uses the axis-wise cycle counts $\hat{r}^{(a)}_{k}$ to decide how much to scale each band for a given extrapolation factor $s_a$.
This makes the positional encoding explicitly training-window aware, band-aware, and AR-aware, and empirically enables more stable 2K/4K multi-AR inference with negligible overhead.


\subsubsection{SNR-Aware Huber Wavelet Training Objective}
\textbf{Motivation.}
Wavelet-space objectives such as \emph{Diffusion4K} demonstrate that measuring errors in a multi-scale transform can materially improve 4K fidelity~\cite{zhang2025diffusion}, but at native 4K we empirically observe that standard $L_2$-based training on VAE latents still suffers from three coupled pathologies. 
(i) \emph{Frequency imbalance}—natural-image wavelet coefficients are heavy-tailed~\cite{wainwright1999scale}, so large high-frequency residuals (textures, edges, micro-geometry) are aggressively shrunk by quadratic losses, leading to over-smoothing of detail. 
(ii) \emph{Timestep imbalance}—gradients concentrate at extremely small or large noise levels, echoing Min-SNR analyses that show inefficient use of intermediate timesteps~\cite{hang2023efficient}. 
(iii) \emph{Cross-scale energy coupling}—low-frequency bands dominate pixel/latent norms, so the high-frequency errors that largely govern 4K perceptual quality receive disproportionately small gradient signal~\cite{zhang2025diffusion}. 
To address these issues, we design a single objective that is simultaneously (a) \emph{robust yet smooth}—using a Pseudo-Huber penalty that behaves like $L_2$ near zero and $L_1$ in the tails~\cite{song2020improved}; (b) \emph{SNR-aware}—with an adaptive threshold $c(t)$ that is small under high noise and grows as signal dominates; (c) \emph{frequency-aware}—by measuring residuals in an orthonormal wavelet space that decouples low and high bands; and (d) \emph{time-rebalanced}—via Min-SNR weighting that emphasizes mid-SNR timesteps for stable, faster optimization~\cite{hang2023efficient}. 
These choices yield the \emph{SNR-Aware Huber Wavelet} objective, a drop-in replacement for standard flow-matching losses tailored to the demands of native 4K generation.

\label{sec:sahw}
\noindent\textbf{Classical FM setup.}
Prior work on flow matching for DiTs adopts a straight-line interpolation between a clean latent $z$ and Gaussian noise $\varepsilon$ \cite{lipman2022flow,batifol2025flux}:
\begin{equation}
z_t=(1-t)\,z+t\,\varepsilon,\qquad t\in(0,1),\ \varepsilon\sim\mathcal N(0,I).
\end{equation}
Under this parameterization, the DiT model predicts a velocity field $v_\theta(z_t,t)$ and the associated data-prediction is
\begin{equation}
\hat z_\theta(z_t,t)=z_t - t\,v_\theta(z_t,t).
\end{equation}

To balance gradients across timesteps, we collapse the straight-path factor and Min-SNR into a single weight
\begin{equation}
\omega(t)=\frac{t}{1-t}\,\min\{\mathrm{SNR}(t),\gamma\}^{\beta}.
\end{equation}
Here $\mathrm{SNR}(t)={(1-t)^2}/{t^2}$ under the straight FM path, with $\gamma>0$ and $\beta\ge0$.
We measure residuals in a wavelet space: letting $\mathcal W(\cdot)$ denote a one-level orthonormal DWT (sub-bands concatenated along channels), we compute the residual \(R_\theta(x,\varepsilon,t)=\mathcal W(\hat z_\theta(z_t,t))-\mathcal W(z)\).
For robustness we use the Pseudo-Huber penalty \(\rho_{c}(r)=c^{2}\!\big(\sqrt{1+(r/c)^{2}}-1\big)\)~\cite{song2020improved} and schedule its threshold as
\begin{equation}
c(t)=c_{\min}+(c_{\max}-c_{\min})\!\left(\frac{\min\{\mathrm{SNR}(t),\gamma\}}{\gamma}\right)^{\alpha}.
\end{equation}
We take $\alpha\in[0,1]$, so that $c(t)$ is small and robust at low SNR and grows smoothly toward high SNR. Let $N$ denote the number of pixels after wavelet stacking.
The per-pixel robust wavelet loss is
\begin{equation}
\label{eq:phuber-wavelet}
\ell_{\text{Huber}}\!\big(R_\theta; c(t)\big)
=\frac{1}{N}\sum_{p=1}^{N}\rho_{\,c(t)}\!\big(R_{\theta,p}\big),
\end{equation}
where $R_{\theta,p}$ is the $p$-th element of $R_\theta(x,\varepsilon,t)$.
Our final objective for the DiT of our UltraFlux becomes
\begin{equation}
\label{eq:sahw}
\mathcal L(\theta)=\mathbb E_{z,\varepsilon,t}\!\Big[
\omega(t)\;\ell_{\text{Huber}}\!\big(R_\theta; c(t)\big)
\Big].
\end{equation}
Setting $c(t)\!\to\!\infty$ and $\beta\!=\!0$ recovers the standard flow-matching objective on this path.

\begin{table}[t!]
\centering
\vspace{-0.6cm}
\captionsetup{font=scriptsize} 
\caption{{Quantitative comparison under 4K res. with open-source methods.}}
\vspace{-0.3cm}
\label{tab:quantitative_comparison_open}
\begin{adjustbox}{max width=\linewidth}
\begin{tabular}{@{}l *{7}{c}@{}}
\toprule
Method & FID $\downarrow$ & HPSv3 $\uparrow$ & PickScore $\uparrow$ & ArtiMuse $\uparrow$ & CLIP Score $\uparrow$ & Q-Align $\uparrow$ & MUSIQ $\uparrow$ \\
\midrule
ScaleCrafter~\cite{he2023scalecrafter}    
    & 164.02 & 6.83   & 21.68 & 67.88 & 33.36 & 4.30 & 38.21 \\
FouriScale~\cite{huang2024fouriscale}      
    & 164.71 & 11.19  & 21.86 & 65.87 & 33.11 & 4.50 & 38.96 \\
Sana~\cite{xie2024sana}                    
    & 144.17 & 10.83  & \textbf{23.18} & 63.72 & \textbf{35.49} & \textbf{4.89} & 45.08 \\
Diffusion-4K~\cite{zhang2025diffusion}     
    & 152.43 & 8.92   & 21.88 & 63.76 & 33.00 & 4.69 & 27.51 \\
\midrule
UltraFlux                                     
    & \textbf{143.11} & \textbf{11.47} & 22.69 & \textbf{68.36} & 34.62 & 4.85 & \textbf{46.13} \\
\bottomrule
\end{tabular}
\end{adjustbox}
\end{table}

\begin{table}[t!]
\centering
\vspace{-0.3cm}
\begin{minipage}{0.48\linewidth}
\captionsetup{font=scriptsize} 
\centering
\caption{Quantitative comparison with Sana at 4096$\times$2048 (2:1) and 2048$\times$4096 (1:2) resolutions.}
\label{tab:sana_multi_ar}
\vspace{-0.35cm}
\begin{adjustbox}{max width=\linewidth}
\begin{tabular}{@{}l *{4}{c}@{}}
\toprule
Method & FID $\downarrow$ & HPSv3 $\uparrow$ & Artimuse $\uparrow$ & Q-Align $\uparrow$ \\
\midrule
Sana (2:1)    
    & 150.35 & 9.01 & 63.61 & 4.80 \\
UltraFlux (2:1)    
    & \textbf{147.53} & \textbf{9.91} & \textbf{64.81} & \textbf{4.86} \\
\midrule
Sana (1:2)    
    & 149.41 & 11.40 & \textbf{66.95} & 4.85 \\
UltraFlux (1:2)    
    & \textbf{143.71} & \textbf{12.51} & 66.41 & \textbf{4.89} \\
\bottomrule
\end{tabular}
\end{adjustbox}
\end{minipage}
\hspace{0.1cm} 
\begin{minipage}{0.48\linewidth}
\centering
\captionsetup{font=scriptsize} 
\caption{Quantitative comparison with Sana at 5120$\times$2880 (16:9) and 5952$\times$2496 (2.39:1) resolutions.}
\vspace{-0.35cm}
\begin{adjustbox}{max width=\linewidth}
\begin{tabular}{@{}l *{4}{c}@{}}
\toprule
Method & FID $\downarrow$ & HPSv3 $\uparrow$ & Artimuse $\uparrow$ & Q-Align $\uparrow$ \\
\midrule
Sana (16:9)
    & 153.31 & 9.04 & 63.02 & 4.81 \\
UltraFlux (16:9)
    & \textbf{142.43} & \textbf{9.91} & \textbf{67.22} & \textbf{4.85} \\
\midrule
Sana (2.39:1)
    & 153.10 & 8.57 & 62.48 & 4.77 \\
UltraFlux (2.39:1)    
    & \textbf{151.98} & \textbf{11.76} & \textbf{66.36} & \textbf{4.82} \\
\bottomrule
\end{tabular}
\end{adjustbox}
\end{minipage}
\end{table}

\begin{table}[t!]
\centering
\vspace{-0.3cm}
\captionsetup{font=scriptsize} 
\caption{{Quantitative comparison under 4K res. with close-source method.}}
\vspace{-0.3cm}
\label{tab:quantitative_comparison_close}
\begin{adjustbox}{max width=\linewidth}
\begin{tabular}{@{}l *{7}{c}@{}}
\toprule
Method & FID $\downarrow$ & HPSv3 $\uparrow$ & PickScore $\uparrow$ & ArtiMuse $\uparrow$ & CLIP Score $\uparrow$ & Q-Align $\uparrow$ & MUSIQ $\uparrow$ \\
\midrule
Seedream 4.0~\cite{seedream2025seedream}    
    & \textbf{132.87} & {11.98} & \textbf{23.52} & \textbf{69.83} & \textbf{35.26} & 4.71 & 30.21 \\
UltraFlux w. Prompt Refiner                                      
    & {147.06} & \textbf{12.03} & 23.25 & {68.75} & 34.50 & \textbf{4.93} & \textbf{45.93} \\
\bottomrule
\end{tabular}
\end{adjustbox}
\vspace{-0.4cm}
\end{table}

\subsubsection{Stage-wise Aesthetic Curriculum Learning}
Recent analyses highlight that diffusion timesteps correspond to qualitatively different generation tasks, with high-noise steps shaping global structure and low-noise steps refining local details~\cite{yi2024diffusion,kim2024denoising}
. Building on this view, OmniSync~\cite{peng2025omnisync} assigns different datasets to different timestep ranges for lip-synchronization, and several works propose timestep curricula or timestep-dependent adaptation to focus training on harder noise regimes~\cite{xu2024towards,soboleva2025t}. In parallel, aesthetic filtering and post-training on high-quality subsets (e.g., LAION-Aesthetics~\cite{schuhmann2022laionaesthetics} and subsequent aesthetic post-training methods~\cite{liang2025aesthetic}) have proven effective for improving visual appeal, but they typically apply a static high-aesthetic prior uniformly across all timesteps.

We instead couple the noise and data axes in a simple two-stage scheme, which we term \emph{Stage-wise Aesthetic Curriculum Learning (SACL)}. In Stage~1, we fine-tune the model on the full \emph{MultiAspect-4K-1M} corpus with standard timestep sampling over the entire diffusion horizon, giving the DiT backbone broad coverage of aspect ratios, content types, and noise levels. Stage~2 then restricts training to a high-noise band—timesteps above a threshold where the model relies most on its generative prior—and to the top-$5\%$ images ranked by the ArtiMuse aesthetic score. Intuitively, Stage~1 learns a general 4K prior across all denoising tasks, while Stage~2 concentrates the remaining compute on the hardest regime with ultra high-aesthetic supervision. Unlike prior timestep curricula, which modulate timestep sampling under a fixed data distribution~\cite{yi2024diffusion,xu2024towards,kim2024denoising}, or aesthetic post-training, which ignores the different roles of noise levels~\cite{liang2025aesthetic}, SACL uses stage-wise aesthetic filtering to define a curriculum over high-noise timesteps, steering the global 4K generative prior toward high-aesthetic modes precisely where the sampling process is most underdetermined and yielding substantial 4K aesthetic and alignment gains at modest additional training cost (Sec.~\ref{sec:experiments}).

\section{Experiment}
\begin{figure}[t!]
 \vspace{-0.8cm}
    \centering
    \captionsetup{font=footnotesize} 
    \includegraphics[width=1\linewidth]{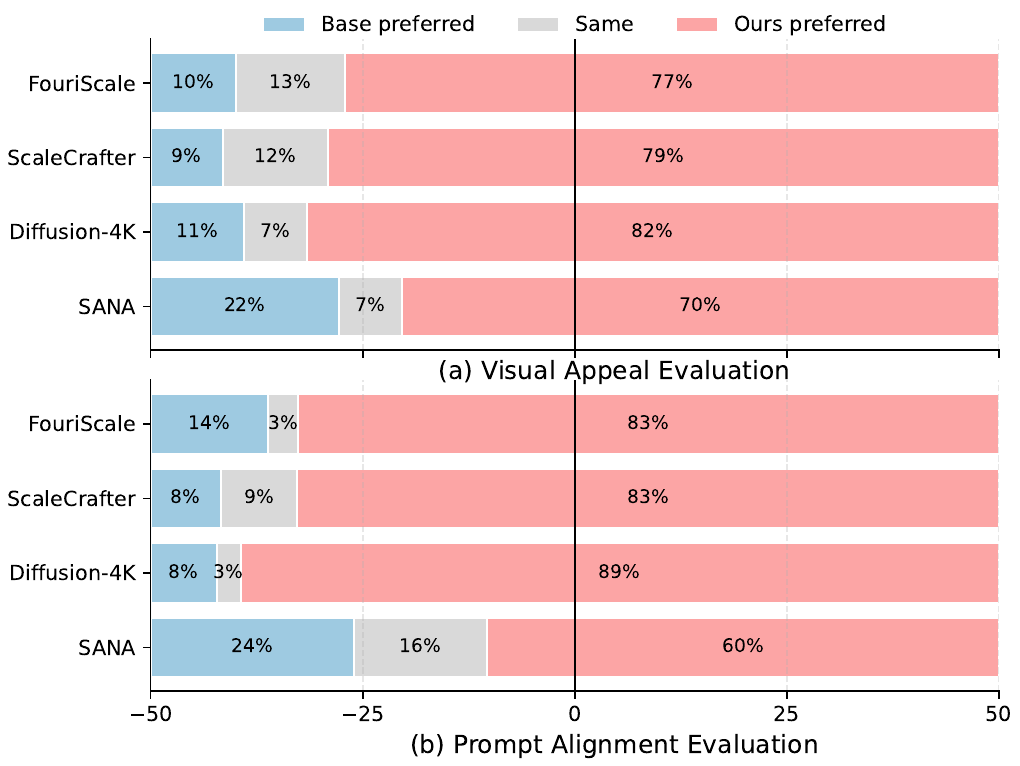}
        \vspace{-0.6cm}
 \caption{{Gemini-2.5-Flash preference comparison.}}
 \vspace{-0.7cm}
    \label{fig:gemini_evaluation}
\end{figure}
\label{sec:experiments}

\begin{figure*}[t!]
\vspace{-0.6cm}
\centering
\includegraphics[width=0.92\linewidth]{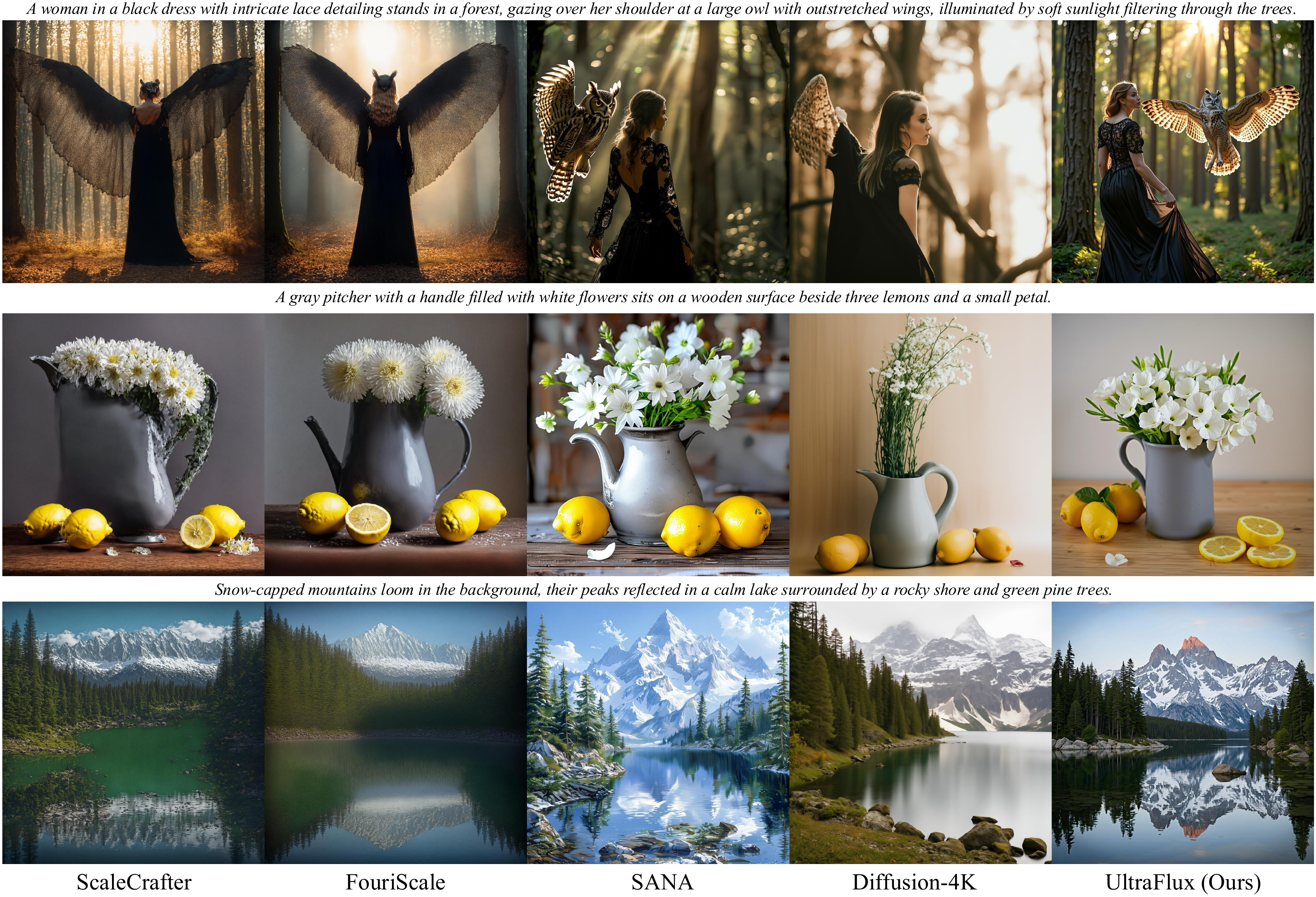}
\vspace{-0.2cm}
  \captionsetup{font=scriptsize} 
\caption{Visual comparison of open-source methods on the Aesthetic-Eval@4096 benchmark at 4096$\times$4096 resolution.}
\label{fig:visual_comparison}
\vspace{-0.6cm}
\end{figure*}

\subsection{Comparison with Open-source Methods.}

\noindent\textbf{Quantitative Comparison.} As shown in Table~\ref{tab:quantitative_comparison_open}, we compare our method against several strong baselines: ScaleCrafter~\cite{he2023scalecrafter} and FouriScale~\cite{huang2024fouriscale} as representative training-free high-resolution scaling methods, Sana~\cite{xie2024sana} as a recent native 4K text-to-image foundation model, and Diffusion-4K~\cite{zhang2025diffusion} as a Flux-based model trained natively at 4K resolution. All methods are evaluated on the Aesthetic-Eval@4096 benchmark~\cite{zhang2025diffusion} using FID~\cite{heusel2017gans}, HPSv3~\cite{ma2025hpsv3widespectrumhumanpreference}, PickScore~\cite{Kirstain2023PickaPicAO}, ArtiMuse~\cite{cao2025artimuse}, CLIP Score~\cite{zhang2024long}, Q-Align~\cite{Q-Align} and MUSIQ~\cite{MUSIQ}.  Beyond the square setting, Table~\ref{tab:sana_multi_ar} further shows that UltraFlux consistently matches or surpasses SANA across a range of non-square 4K aspect ratios, delivering better semantic alignment, and perceptual quality even in challenging panoramic and wide ARs. We also provide visual comparison as show in Figure ~\ref{fig:visual_comparison}. 

\noindent\textbf{Gemini-based Preference Evaluation.}
To complement automatic metrics, we conduct a large-scale pairwise preference study using \emph{Gemini-2.5-Flash} in reasoning mode as an LMM judge. For each prompt and baseline, Gemini is shown the prompt and two anonymized images (baseline vs.\ UltraFlux) and asked which it prefers in terms of \emph{visual appeal} and \emph{prompt alignment}, with ties allowed. As summarized in Figure~\ref{fig:gemini_evaluation}, UltraFlux is preferred in $70$–$82\%$ of cases on visual appeal and $60$–$89\%$ on prompt alignment across all comparisons.
\subsection{Comparison with Close-source 4K Native Generation Models}
As shown in Table.~\ref{tab:quantitative_comparison_close} To make the comparison with \emph{Seedream 4.0}~\cite{seedream2025seedream} as fair as possible, we mirror its use of a large-scale LLM-based prompt refiner by evaluating \emph{UltraFlux w. Prompt Refiner (Ours)} with a GPT-4O front-end. Under the same 4096$\times$4096 evaluation protocol, Table~\ref{tab:quantitative_comparison_close} shows that UltraFlux with GPT-4O achieves a slightly higher HPSv3 score than Seedream 4.0 (12.03 vs.\ 11.98), while remaining competitive on PickScore, ArtiMuse, and CLIP Score, and clearly surpassing it on Q-Align and MUSIQ, which better capture semantic alignment and perceptual image quality. This indicates that, once both systems are equipped with strong prompt refiners, our open-source UltraFlux model trained on 1M images can closely track---and in some aspects exceed—the performance of a leading proprietary 4K generator. \textit{Notably, Seedream~4.0 further benefits from large-scale RL post-training, whereas our full pipeline relies solely on stage-wise SFT.} 
\subsection{Ablation Study}
\begin{table}[t!]
\centering
  \captionsetup{font=scriptsize} 
  \vspace{0cm}
\caption{{
Ablation study on UltraFlux at 4K resolution training.
}}
\label{tab:ultraflux_ablation}
  \vspace{-0.2cm}
\begin{adjustbox}{max width=1\linewidth}
\begin{tabular}{@{}lccc@{}}
\toprule
Variant & FID $\downarrow$ & HPSv3 $\uparrow$ & ArtiMuse $\uparrow$ \\
\midrule
Flux+F16 VAE (base) & 151.40 & 9.22 & 66.39 \\
+ SNR-HW & 148.81 & 9.70 & 67.23 \\
+ SNR-HW + SACL & 147.32 & 10.30 & 67.31 \\
+ SNR-HW + SACL + Resonance 2D RoPE w.\ YARN & 146.93 & 10.91 & 68.13 \\
\bottomrule
\vspace{-0.5cm}
\end{tabular}
\end{adjustbox}
\end{table}

Starting from a Flux and trained F16 VAE baseline, replacing the standard latent regression loss with our SNR-Aware Huber Wavelet Training (SNR-HW) objective already yields consistent gains across metrics under the same 500K data$\&$10K steps fine-tuning schedule, indicating that SNR-aware wavelet supervision better balances high-frequency detail preservation with stable optimization. Introducing the SACL term on top of SNR-HW further improves both human preference and aesthetic scores, suggesting that stronger text–image alignment is especially beneficial at native 4K. Finally, equipping UltraFlux with Resonance 2D RoPE and YARN produces the best overall configuration, delivering monotonically improved perceptual and aesthetic metrics while also reducing FID. Taken together, these ablations show that the proposed objective, alignment loss, and positional encoding contribute complementary gains rather than merely redistributing performance among metrics.

\section{Conclusion}
UltraFlux couples a carefully curated MultiAspect-4K-1M corpus with AR-aware positional encoding, reconstruction, and optimization components into a unified framework for native 4K multi-AR generation. This data--model co-design yields state-of-the-art fidelity, aesthetic quality, and text alignment on standard 4K benchmarks while remaining computationally practical.

{
\small
\bibliographystyle{ieeenat_fullname}
\bibliography{main}
}

\clearpage
\setcounter{page}{1}
\maketitlesupplementary

\section*{Supplementary Overview}
This document complements the main paper by (i) clarifying the regime-level novelty of \emph{UltraFlux} as a data--model co-designed system for native 4K, multi-AR text-to-image generation, and (ii) providing the analyses, metrics, and implementation details needed to faithfully reproduce and stress-test our setup. Rather than introducing isolated primitives, we make explicit how dataset curation, positional representation, VAE compression, and optimization objectives must be co-designed to operate robustly in the 4K, multi-AR regime.

\noindent\textbf{Organization.}
Sec.~S1 (\emph{Clarifying Novelty and Data--Model Co-Design}) positions UltraFlux as a regime-level, system-oriented contribution and introduces a 2$\times$2 data--model ablation that disentangles the roles of \textsc{MultiAspect-4K-1M} and the UltraFlux architecture/loss.  
Sec.~S2 (\emph{Implementation Details}) expands on the dataset pipeline, DiT training schedule, VAE post-training recipe, and key hyperparameters, including reconstruction metrics for our F16 VAE.  
Sec.~S3 and Sec.~S4 provide focused analyses of Resonance 2D RoPE with YaRN and wavelet-space statistics of 4K VAE latents, together with qualitative 4K visualizations, motivating our positional design and SNR-Aware Huber Wavelet objective.  
Sec.~S5–S7 report additional ablations, 4K runtime measurements, and more extensive quantitative comparisons at challenging wide aspect ratios, as well as extended visual comparisons against open-source baselines.  
Sec.~S8 discusses the main limitations of UltraFlux in terms of sampling cost, memory footprint, and aesthetic ceiling, while Sec.~S9 details our Gemini-based preference evaluation and GPT-4o prompt-refiner setup used for large-scale automatic assessment and prompt expansion.

Taken together, these sections are intended to show that UltraFlux is a \emph{\textbf{regime-level, system-oriented contribution}} rather than a mere aggregation of existing tricks, and to document the concrete choices required to make native-4K, multi-AR generation work in practice.

\section{Clarifying Novelty and Data–Model Co-Design}
The main paper positions UltraFlux as a \textbf{\emph{data–model co-designed recipe}} for native-4K, multi-AR text-to-image generation. Several of the building blocks—resonance-style rotary encodings, wavelet objectives, Min-SNR weighting, and aesthetic curricula—indeed draw inspiration from prior work. Our contribution is not to claim each primitive as a standalone invention, but to show that: (i) at 4K with diverse aspect ratios, positional encoding, VAE compression, and optimization objectives form a \textbf{\emph{coupled regime}} that existing methods treat largely in isolation; and (ii) a carefully unified design across \textbf{\emph{dataset, representation, and loss}} yields behaviors that cannot be reproduced by swapping in any single component in isolation.

To make this clearer, we provide in this supplement:
\begin{itemize}

  \item A data-model ablation (Table~\ref{tab:s0_co_design_ablation}) showing that neither a stronger 4K dataset nor architectural changes alone are sufficient: MultiAspect-4K-1M and UltraFlux each yield modest gains in isolation, while their combination delivers the full non-additive improvements in 4K, multi-AR fidelity.

  \item One-dimensional and two-dimensional diagnostics of Resonance 2D RoPE with YaRN (Sec.~\ref{subsec:Analyses-RoPE}), analyzing cycle snapping, phase closure on the training window, and the stability of phase geometry under aspect-ratio extrapolation.

  \item Wavelet-space statistics of 4K VAE latents (Sec.~\ref{subsec:wavelet-stats}) that empirically confirm the low-frequency–dominated yet heavy-tailed structure motivating our SNR-Aware Huber Wavelet objective, clarifying why a robust, SNR-aware wavelet loss is better aligned with the 4K regime than a pure latent $L_2$ objective.

  \item Expanded implementation details for the dataset pipeline, DiT training, and VAE post-training (Sec.~S2), to facilitate faithful reproduction of our 4K native, multi-AR training setup.

\end{itemize}

We hope these analyses better convey that UltraFlux is a \emph{\textbf{regime-level, system-oriented contribution}} rather than a mere aggregation of existing tricks.

\begin{table}[t]
\centering
\caption{2$\times$2 data--model co-design ablation. 
A: baseline; B: data only; C: model/loss only; D: full co-design.}
\label{tab:s0_co_design_ablation}
\scriptsize
\begin{tabular}{@{}c l l c c@{}}
\toprule
Variant & Dataset & Model / Loss & FID $\downarrow$ & HPSv3 $\uparrow$ \\
\midrule
A & Diffusion-4K-v2~\cite{zhang2025ultra}           & Flux, latent L2                                     & 152.09  & 8.57 \\
B & MultiAspect-4K-1M & Flux, latent L2                                      & 151.41  & 9.17 \\
C & Diffusion-4K-v2~\cite{zhang2025ultra}             & UltraFlux   &147.41  &10.03  \\
D & MultiAspect-4K-1M & UltraFlux   &145.81  &10.78  \\
\bottomrule
\end{tabular}
\end{table}

\section{Implementation Details}
\subsection{Dataset Pipeline}
\noindent \textbf{Flat-region detection.}  
For each image, we first partition it into non-overlapping \(240 \times 240\) patches and quantify the edge richness of every patch with a Sobel-based score,
\[
S_{\text{flat}} = \mathrm{Var}\!\left(\sqrt{(\partial_x I)^2 + (\partial_y I)^2}\right).
\]
Patches with \(S_{\text{flat}} < 800\) are flagged as texture-poor, and any image in which more than \(50\%\) of the patches are flagged is removed from the dataset. The patch-level threshold of \(800\) and the \(50\%\) image-level ratio are selected empirically via manual inspection of edge-statistic histograms and visual audits. This conservative configuration effectively filters out images dominated by large uniform regions while still retaining plausible low-texture content such as sky and water, ensuring that the remaining images maintain sufficient edge and texture diversity for high-fidelity generation.

\noindent \textbf{Information Entropy Filtering.}  
Each image is analyzed for its Shannon entropy to quantify the amount of information it contains. The Shannon entropy \(H\) of an image is defined as:
\[
H = - \sum_{i=1}^{N} p(x_i) \log_2 p(x_i),
\]
where \(p(x_i)\) denotes the probability of the pixel value \(x_i\) within the image. Images with an entropy value \(H < 7.0\) are flagged as texture-poor, and any image in which \(H < 7.0\) is removed from the dataset. The threshold of \(7.0\) is selected empirically based on the observed distribution of entropy values across the dataset. This threshold effectively filters out images with insufficient texture or information, ensuring that the remaining images exhibit adequate variability for high-quality processing while preserving content diversity.

\noindent\textbf{Image Quality Filtering.}  
To ensure semantic quality, we compute the quality score for each image using \emph{Q-Align}~\cite{wu2023qalign}. Images with a quality score greater than \(4.0\) are retained, while those below this threshold are discarded. This threshold is determined empirically based on the distribution of quality scores across data sources, ensuring that only images with sufficient semantic clarity are kept for further analysis.

\noindent\textbf{Aesthetic Quality Filtering.}  
For aesthetic evaluation, we use the \emph{ArtiMuse}~\cite{cao2025artimusefinegrainedimageaesthetics} model to compute aesthetic scores for each image. Only the top \(30\%\) of images, based on their aesthetic rating, are preserved. This strategy ensures that images with higher aesthetic appeal are prioritized, while lower-rated images are excluded from the dataset. This filtering method helps maintain a diverse and aesthetically pleasing selection of images for further processing.

\subsection{Training Details}
\noindent\textbf{DiT Training.} We train \emph{UltraFlux}, a large Flux-based DiT model for native 4K text-to-image generation. During DiT training, we freeze the VAE and text encoders and update all DiT blocks end-to-end. Training is conducted on $8\times$NVIDIA H800 GPUs using DeepSpeed ZeRO-2 (without CPU offload). We choose ZeRO-2 because it shards optimizer states and gradients without partitioning model parameters, which substantially reduces memory usage while yielding higher throughput than ZeRO-3 in our setting, enabling efficient 4K training. We use AdamW with a learning rate of $1\times 10^{-6}$ and an effective batch size of 64; the full training run takes roughly 12 days. We adopt a two-stage training schedule, with approximately 30K steps in the first stage and a further 2K steps in the second fine-tuning stage (Stage-wise Aesthetic Curriculum Learning). To support multi-AR native 4K generation, we adopt a bucketed resolution scheme: for each image, we snap its resolution to the nearest target from a fixed set of landscape buckets (e.g., $5120\times 2880$ for 16:9, $4704\times 3136$ for 3:2), portrait buckets (e.g., $2880\times 5120$ for 9:16, $3136\times 4704$ for 2:3), and a single square bucket at $3840\times 3840$, then center-crop and resize the image to the selected bucket resolution.

\noindent\textbf{VAE Training.} For VAE post-training, we fine-tune the decoder on the proposed \emph{MultiAspect-4K-1M} dataset, retaining the top $50\%$ of images according to the flatness score and training at $512\times512$ resolution with an effective batch size of 384. We use AdamW with a learning rate of $1\times10^{-5}$.

\begin{table}[t]
\centering
\caption{Reconstruction metrics of F16 VAEs on the Aesthetic-4K@4096 Eval set~\cite{zhang2025diffusion}.}
\label{tab:vae_recon_metrics}
\scriptsize
\begin{adjustbox}{width=1\linewidth}
\begin{tabular}{l|ccccc}
\toprule
Model & rFID $\downarrow$ & NMSE $\downarrow$ & PSNR $\uparrow$ & SSIM $\uparrow$ & LPIPS $\downarrow$ \\
\midrule
Flux-VAE-F16~\cite{zhang2025diffusion}       & 2.201 & 0.01522 & 26.90 & 0.784 & 0.168 \\
Flux-VAE-F16-SC~\cite{zhang2025ultra}       & 0.588 & 0.00736 & 30.19 & 0.846 & 0.097 \\
\textbf{UltraFlux-F16-VAE} & \textbf{0.547} & \textbf{0.00657} & \textbf{30.70} & \textbf{0.852} & \textbf{0.102} \\
\bottomrule
\end{tabular}
\end{adjustbox}
\end{table}

\noindent\textbf{VAE reconstruction metrics and post-training gains.}
Table~\ref{tab:vae_recon_metrics} quantitatively compares our \textbf{UltraFlux-F16-VAE} with the Flux-VAE-F16 baseline on the \textsc{Aesthetic-4K@4096} evaluation set~\cite{zhang2025diffusion}. Despite using the same F16 compression ratio, UltraFlux-F16-VAE achieves substantially better reconstruction quality across all metrics. These consistent gains indicate that our post-trained decoder not only preserves low-frequency structure, but also better reconstructs high-frequency details that are typically washed out under aggressive F16 compression. Combined with the wavelet-space analysis above, this suggests that the proposed post-training scheme effectively aligns the VAE with the heavy-tailed, cross-scale statistics of native 4K images, narrowing the reconstruction gap.


\begin{table}[t]
\centering
\caption{Summary of training-related hyperparameters for UltraFlux and associated components. Values are left blank to be filled with the final configuration.}
\label{tab:training_hparams}
\scriptsize
\begin{adjustbox}{max width=\linewidth}
\begin{tabular}{llc}
\toprule
Component & Hyperparameter & Value \\
\midrule
\multirow{3}{*}{Stage-wise Aesthetic Curriculum}
  & Stage~1 timestep range & 0--999 \\[2pt]
  & Stage~2 timestep range & 0--459 \\[2pt]
  & Stage~2 aesthetic filter (ArtiMuse percentile) & top-$5\%$ \\[2pt]
\midrule
\multirow{2}{*}{DiT objective}
  & Wavelet type / number of levels & Haar, $J{=}1$ \\[2pt]
  & Pseudo-Huber thresholds $(c_{\min},c_{\max})$ & $c_{\min}{\approx}0.2,\;c_{\max}{\approx}1.0$ \\[2pt]
\midrule
\multirow{4}{*}{Resonance 2D RoPE with YaRN}
  & RoPE base $b$ & $10{,}000$ \\[2pt]
  & NTK scaling factor $\eta$ & $1.0$ \\[2pt]
  & YaRN ramp parameters $(\alpha,\beta)$ & $(1.25,\,0.75)$ \\[2pt]
  & Maximum extrapolation scale $s_a=L'_a/L_a$ & $2.0$ \\[2pt]
\midrule
\multirow{4}{*}{F16 VAE post-training}
  & Training resolution & $512 \times 512$ \\[2pt]
  & Global batch size (images/step) & 384 \\[2pt]
  & Optimizer / learning rate / weight decay & AdamW, $1\times10^{-4}$, $1\times10^{-2}$ \\[2pt]
  & Loss weights $(\lambda_{\text{wav}},\lambda_{\text{perc}},\lambda_{L_2})$ & $0.2,\,0.1,\,1$ \\[2pt]
\midrule
\multirow{7}{*}{Multi-AR 4K DiT training}
  & Landscape target sizes (W$\times$H) 
  & $5440{\times}3072$, $5184{\times}3264$, $4992{\times}3328$ \\[2pt]
  & & $4736{\times}3520$, $5824{\times}2880$, $6272{\times}2688$ \\[2pt]
  & & $5568{\times}3008$, $6336{\times}2624$, $5632{\times}3008$ \\[2pt]
  & & $4608{\times}3648$ \\[2pt]
  & Portrait target sizes (W$\times$H) 
  & $3072{\times}5440$, $3648{\times}4608$, $3520{\times}4736$ \\[2pt]
  & & $3328{\times}4992$ \\[2pt]
  & Square target sizes (W$\times$H) 
  & $4096{\times}4096$ \\[2pt]
\bottomrule
\end{tabular}
\end{adjustbox}
\end{table}

\begin{figure*}[htbp!]
    \centering
    \includegraphics[width=0.98\linewidth]{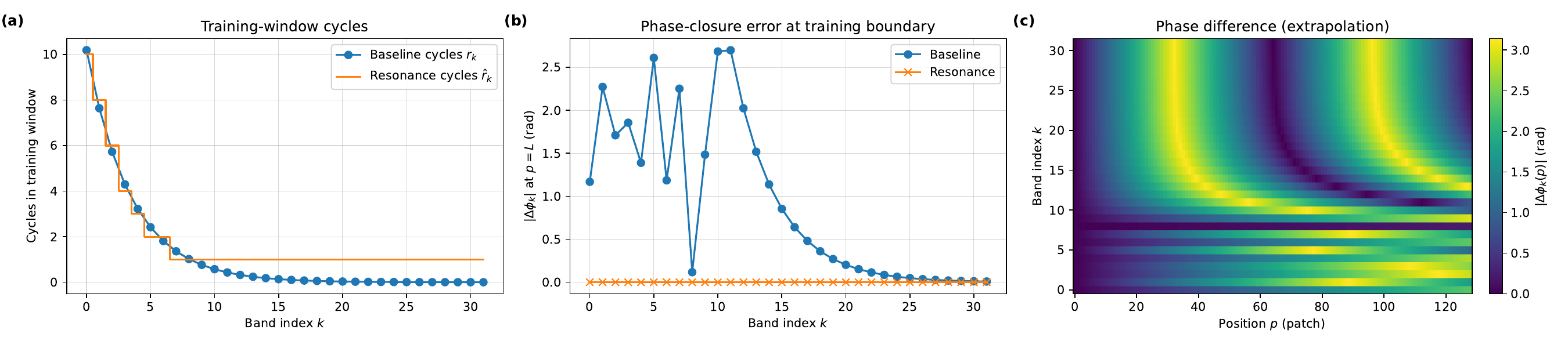}
\caption{
1D band-wise analysis of Resonance 2D RoPE with YaRN.  
(a) Number of cycles $r_k$ in the training window and their integer-snapped counterparts $\hat r_k$.  
(b) Phase-closure error $|\Delta\phi_k|$ at $p=L$, showing exact closure for Resonance RoPE.  
(c) Phase difference $|\Delta\phi_k(p)|$ between baseline and Resonance under $2\times$ length extrapolation, illustrating how fractional cycles in the baseline accumulate into large out-of-distribution phase errors.
}

    \label{fig:resonance-1d-analysis}
\end{figure*}

\begin{figure*}[htbp!]
    \centering
    \includegraphics[width=0.98\linewidth]{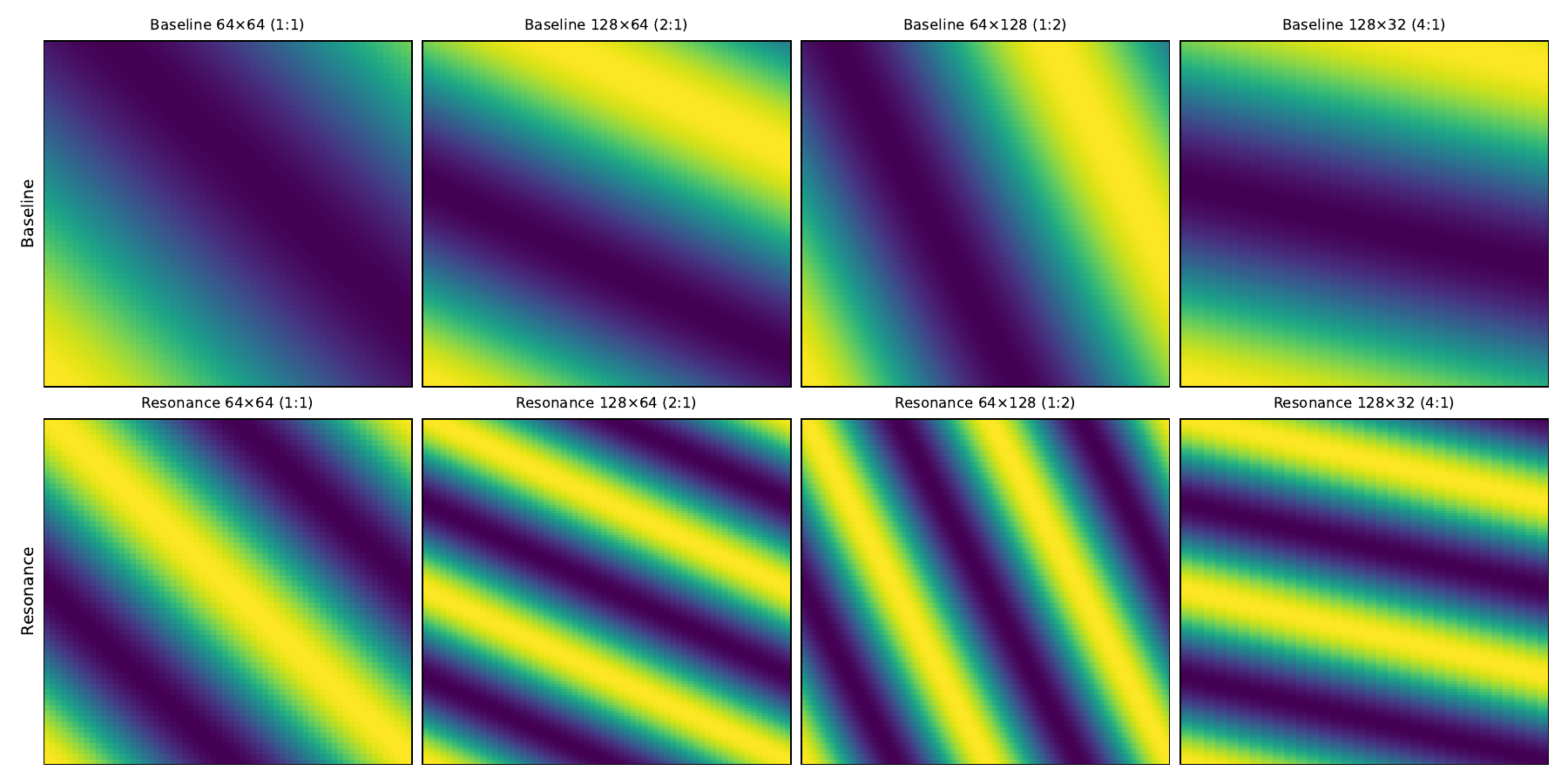}
    \caption{Height--width cosine phase patterns for a representative rotary band under different aspect ratios.  
Each panel displays $f(h,w)=\cos(h\,\omega_H + w\,\omega_W)$ evaluated on an $H{\times}W$ patch grid, with the top row using the Flux baseline frequencies and the bottom row using our resonant frequencies.  
The first column shows the training resolution ($64{\times}64$ patches, AR $1{:}1$); the remaining columns visualize extrapolation to $128{\times}64$ ($2{:}1$), $64{\times}128$ ($1{:}2$), and $128{\times}32$ ($4{:}1$).  
At train scale, baseline and Resonance RoPE induce similar diagonal stripe patterns, indicating that resonance acts as a mild reparameterization of the spectrum.  
Under aspect-ratio extrapolation, the baseline stripes rotate and change spacing more aggressively, while our resonant variant maintains more regular, coherent patterns along both height and width, qualitatively echoing the improved phase stability observed in the 1D analyses.}
    \label{fig:resonance-2d-analysis}
\end{figure*}

\section{Analyses of Resonance 2D RoPE with YaRN}
\label{subsec:Analyses-RoPE}
Figure~\ref{fig:resonance-1d-analysis} gives a 1D band-wise diagnostic of Resonance 2D RoPE on a single spatial axis, which is then used by YaRN.  
In panel (a), we plot the cycles completed in the training window of length $L_a$ by each rotary band,
\[
r^{(a)}_{k} \;=\; \frac{L_a\,\omega^{(a)}_{k}}{2\pi},
\]
together with their snapped counterparts
\[
\hat{r}^{(a)}_{k} \;=\; \max\!\bigl(1,\;\mathrm{round}(r^{(a)}_{k})\bigr).
\]
The Flux baseline (blue) yields a dense sequence of non-integer $r^{(a)}_{k}$, whereas Resonance RoPE (orange) projects every band onto the nearest nonzero integer $\hat{r}^{(a)}_{k}$, producing a piecewise-constant spectrum that leaves low-frequency modes almost unchanged and regularizes higher ones.

Panel (b) measures phase closure at the boundary of the training window.  
For each band we evaluate the phase at $p_a=L_a$ using both the original frequency $\omega^{(a)}_{k}$ and the resonant frequency
\[
\hat{\omega}^{(a)}_{k} \;=\; \frac{2\pi\,\hat{r}^{(a)}_{k}}{L_a},
\]
and plot the absolute phase mismatch $|\Delta\phi_k|$ between $p_a=0$ and $p_a=L_a$.  
The baseline shows up to several radians of mismatch, while Resonance RoPE drives $|\Delta\phi_k|$ to zero for all bands, confirming that every component becomes an exact standing wave on $[0,L_a]$.

Panel (c) visualizes the phase difference between the baseline and Resonance RoPE under a $2\times$ resolution extrapolation.  
For positions $p_a\in[0,2L_a]$ we compute
\[
\Delta\phi_k(p_a) \;=\; \mathrm{wrap}\bigl(p_a\,\omega^{(a)}_{k} - p_a\,\hat{\omega}^{(a)}_{k}\bigr),
\]
where $\mathrm{wrap}(\cdot)$ maps angles to $[-\pi,\pi]$, and plot $|\Delta\phi_k(p_a)|$ as a heatmap over $(k,p_a)$.  
The discrepancy is small near the training window but grows systematically with both position and frequency, illustrating how fractional cycles in the original spectrum accumulate into large out-of-distribution phase errors.  
Since YaRN subsequently applies band-wise scaling to these already integer-cycle–aligned modes, the combined Resonance 2D RoPE with YaRN inherits training-window awareness while achieving stable, AR-robust extrapolation in 2D.

\vspace{0.2cm}
\noindent\textbf{2D spatial visualization.}
Figure~\ref{fig:resonance-1d-analysis} analyzes Resonance 2D RoPE with YaRN along a single spatial axis.  
To understand how these band-wise changes translate into actual image-plane geometry, we further visualize 2D cosine patterns in Figure~\ref{fig:resonance-2d-analysis}.  
For a representative rotary band, we construct
\[
f(h,w) \;=\; \cos\!\big(h\,\omega_H + w\,\omega_W\big),
\]
on different height–width grids, where $(\omega_H,\omega_W)$ are taken either from the Flux baseline or from the resonant frequencies.  
The leftmost column corresponds to the training resolution ($64{\times}64$ patches, AR $1{:}1$), while the remaining columns show extrapolation to $128{\times}64$ ($2{:}1$), $64{\times}128$ ($1{:}2$), and $128{\times}32$ ($4{:}1$).  
At the training scale, baseline and Resonance RoPE produce very similar diagonal stripe patterns, consistent with the fact that snapping $r_k$ to $\hat r_k$ only slightly perturbs low-frequency modes.  
Across more extreme aspect ratios, however, the baseline stripes exhibit more pronounced changes in orientation and spacing, whereas the Resonance patterns remain more regular and coherent.  
This 2D view complements the 1D diagnostics: once each band forms an integer-cycle standing wave on the training window, spatial phase geometry varies more smoothly when scaling to multi-AR 2K/4K grids.

\section{Wavelet-Space Statistics of 4K VAE Latents}
\label{subsec:wavelet-stats}
\begin{figure}[htbp!]
    \centering
    \includegraphics[width=1\linewidth]{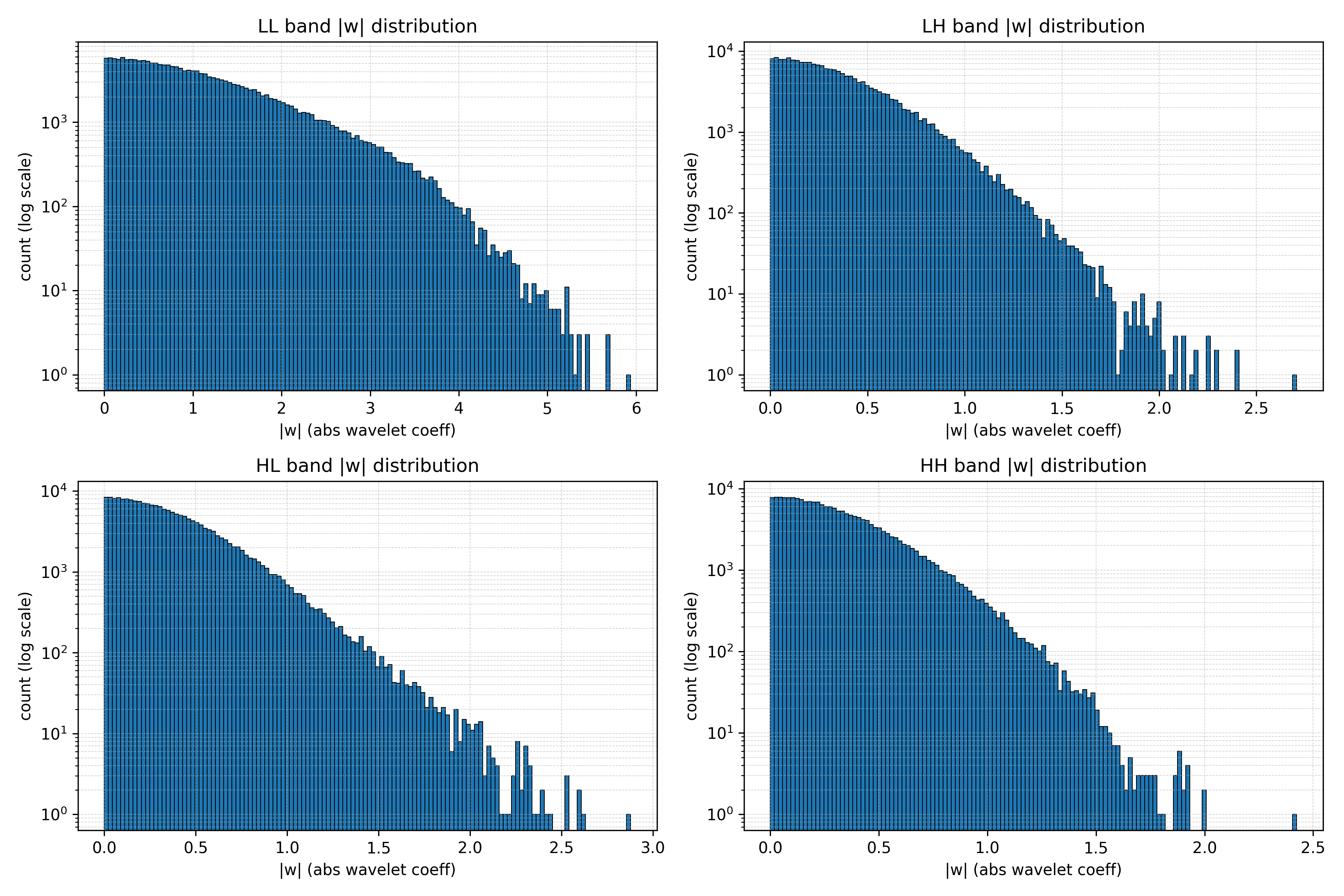}
    \caption{Wavelet-space statistics of 4K VAE latents. We show log-count histograms of absolute coefficients for the LL, LH, HL, and HH subbands over 400 samples. Most energy resides in the LL band, while high-frequency bands carry sparse but large-magnitude coefficients, indicating heavy-tailed behavior. This cross-scale structure motivates our SNR-Aware Huber Wavelet objective.}
    \label{fig:Wavelet-space-statistics}
\end{figure}
In the main paper (Section~3.2.3, \emph{SNR-Aware Huber Wavelet Training Objective}) we argue that native 4K generation suffers from \emph{\textbf{(i) frequency imbalance and (ii) cross-scale energy coupling: low-frequency bands dominate latent norms, while high-frequency, perceptually critical structures appear as sparse, large-magnitude coefficients that are poorly handled by purely quadratic losses. }} Here we provide an empirical characterization of this effect in the VAE latent space used by UltraFlux. We sample 400 images from \textsc{MultiAspect-4K-1M}, encode them with our F16 VAE, and apply a one-level orthonormal DWT to the resulting latents. Figure~\ref{fig:Wavelet-space-statistics} shows log-count histograms of the absolute wavelet coefficients in the LL, LH, HL, and HH subbands. The energy distribution is strongly skewed across scales: the LL band accounts for \textbf{87.4\%} of the total latent energy (mean per-band energy $3.55$), while each high-frequency band contributes only \mbox{$3.5$–$4.7\%$}. At the same time, all bands exhibit pronounced heavy tails. For example, in the LH band \mbox{$20.8\%$} of coefficients satisfy $|w|>0.5$, \mbox{$3.2\%$} exceed $|w|>1.0$, and values up to $|w|\approx 7.2$ occur; HL and HH show similar tail behavior.

\begin{figure*}[t]
    \centering
    \includegraphics[width=0.99\linewidth]{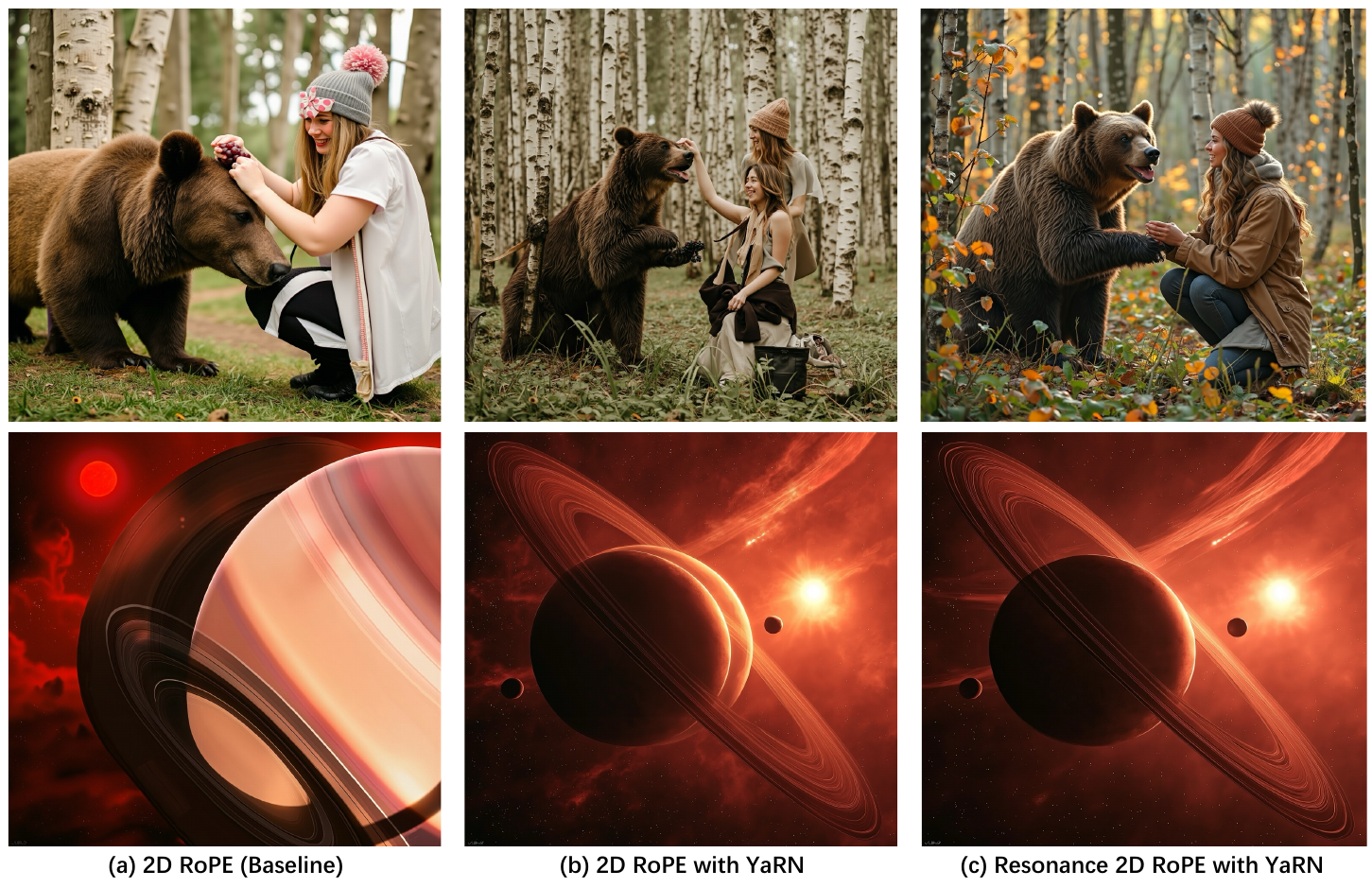}
    \captionsetup{font=scriptsize} 
    \caption{\textbf{Qualitative effect of Resonance 2D RoPE with YaRN.}
    We compare three positional encodings at native 4K resolution for the same prompts. 
    (a) Flux.1 2D RoPE baseline \emph{without} any scaling at inference time, which tends to exhibit geometric drift and mild striping or warping artifacts in both foreground objects and backgrounds. 
    (b) 2D RoPE with YaRN scaling, which stabilizes the overall layout but still shows subtle distortions along long contours and in extreme regions of the image. 
    (c) Our proposed \emph{Resonance 2D RoPE with YaRN}, which yields the most coherent global geometry and sharper, more regular fine structures (e.g., ring edges and tree trunks).}
    \label{fig:ultra_flux_supp_image}
\end{figure*}

These statistics quantitatively support the motivation in the main paper: at 4K resolution, VAE latents are dominated by low-frequency energy, yet contain sparse, large-magnitude high-frequency coefficients that encode textures, edges, and micro-geometry. Under a standard $L_2$ latent loss, these heavy-tailed residuals are aggressively shrunk, and gradients are largely governed by the LL band, leading to over-smoothing of detail and weak supervision for high-frequency errors. This empirical evidence motivates our design of the SNR-Aware Huber Wavelet objective, which replaces pure $L_2$ with a wavelet-space, Pseudo-Huber penalty with SNR-dependent thresholds to better balance low- and high-frequency reconstruction errors in the 4K regime.

\section{Additional Ablations}

Figure~\ref{fig:ultra_flux_supp_image} provides a visual counterpart to the analyses in Sec.~\ref{subsec:Analyses-RoPE}. 
The Flux.1 2D RoPE baseline without scaling (a) reuses its training-time spectrum at 4K and produces noticeable geometric drift: objects appear slightly stretched or misaligned and backgrounds show faint striping.  
Introducing YaRN scaling alone (b) reduces these artifacts by making the spectrum resolution-aware, but residual phase misalignment still leads to mild warping along long contours.  
Our Resonance 2D RoPE with YaRN (c) first snaps each band to an integer-cycle standing wave on the training window and then applies band-wise YaRN scaling, yielding visibly more stable composition and cleaner high-frequency details, especially in the delicate structures of fur, foliage, and planetary rings.

\section{Efficiency Comparison}
\begin{table}[t]
\centering
\caption{Inference time per 4K sample at 4096$\times$4096 resolution.}
\label{tab:4k_infer_time}
\begin{adjustbox}{width=0.9\linewidth
}
\begin{tabular}{lcccc}
\toprule
 & ScaleCrafter & FouriScale & Sana & UltraFlux \\
\midrule
Time (s) & 195.67 & 216.27 & 48.42 & 49.50 \\
\bottomrule
\end{tabular}
\end{adjustbox}
\end{table}

Table~\ref{tab:4k_infer_time} reports the wall-clock time required for each method to generate a single 4096$\times$4096 sample under the same hardware and sampler configuration.
UltraFlux and Sana operate in a similar runtime regime, while both are several times faster than ScaleCrafter and FouriScale, whose 4K pipelines incur substantially higher latency.
In other words, UltraFlux achieves our best 4K fidelity and aesthetic metrics without introducing extra inference cost relative to the strongest open baseline, and remains markedly more efficient than earlier 4K upsampling-based approaches.

\section{More Quantitative Comparison with SOTA Methods at Wide Aspect Ratios}
To provide a more comprehensive evaluation of performance at challenging wide aspect ratios, including 2:1 (4096×2048), 1:2 (2048×4096), 16:9 (5120×2880), and the cinematic 2.39:1, we compare with SOTA methods across four distinct acpect ratios and resolutions, as detailed in Table~\ref{tab:more_multi_ar}. The results show that UltraFlux consistently surpasses the performance all competing methods across all tested aspect ratios and metrics, demonstrating its effectiveness in generating high-quality images for diverse wide-format scenarios.

\begin{table}[t!]
\centering
\caption{Quantitative comparison with SOTA methods at different aspect ratios, including 4096$\times$2048 (2:1), 2048$\times$4096 (1:2), 5120$\times$2880 (16:9), and 5952$\times$2496 (2.39:1) resolutions.}
\label{tab:more_multi_ar}
\begin{adjustbox}{max width=0.98\linewidth}
\begin{tabular}{@{}lc*{4}{c}@{}}
\toprule
Aspect Ratio & Method & FID $\downarrow$ & HPSv3 $\uparrow$ & Artimuse $\uparrow$ & Q-Align $\uparrow$ \\
\midrule
\multirow{4}{*}{\textbf{2:1}} & ScaleCrafter & 168.29 & 6.26 & 65.62 & 4.29 \\
                             & FouriScale & 169.30 & 5.89 & 64.29 & 4.38 \\
                             & Sana & 150.36 & 9.01 & 63.61 & 4.81 \\
                             & UltraFlux & \textbf{147.54} & \textbf{9.91} & \textbf{64.81} & \textbf{4.86} \\
\midrule
\multirow{4}{*}{\textbf{1:2}} & ScaleCrafter & 157.21 & 8.92 & 68.74 & 4.41 \\
                             & FouriScale & 159.87 & 8.09 & 66.64 & 4.38 \\
                             & Sana & 149.42 & 11.40 & \textbf{66.95} & 4.86 \\
                             & UltraFlux & \textbf{143.71} & \textbf{12.51} & 66.41 & \textbf{4.89} \\
\midrule
\multirow{4}{*}{\textbf{16:9}} & ScaleCrafter & 175.97 & 5.30 & 65.05 & 4.14 \\
                              & FouriScale & 173.84 & 5.46 & 64.14 & 4.36 \\
                              & Sana & 153.31 & 9.04 & 63.02 & 4.82 \\
                              & UltraFlux & \textbf{142.43} & \textbf{9.92} & \textbf{67.22} & \textbf{4.85} \\
\midrule
\multirow{4}{*}{\textbf{2.39:1}} & ScaleCrafter & 196.60 & 3.69 & 64.02 & 4.00 \\
                                & FouriScale & 196.30 & 3.61 & 63.09 & 4.14 \\
                                & Sana & 153.10 & 8.57 & 62.48 & 4.77 \\
                                & UltraFlux & \textbf{151.99} & \textbf{11.76} & \textbf{66.36} & \textbf{4.82} \\
\bottomrule
\end{tabular}
\end{adjustbox}
\end{table}

\section{Qualitative Comparison with SOTA Methods at Wide Aspect Ratios}
\begin{figure*}[t!]
\centering
\includegraphics[page=1, width=0.92\linewidth]{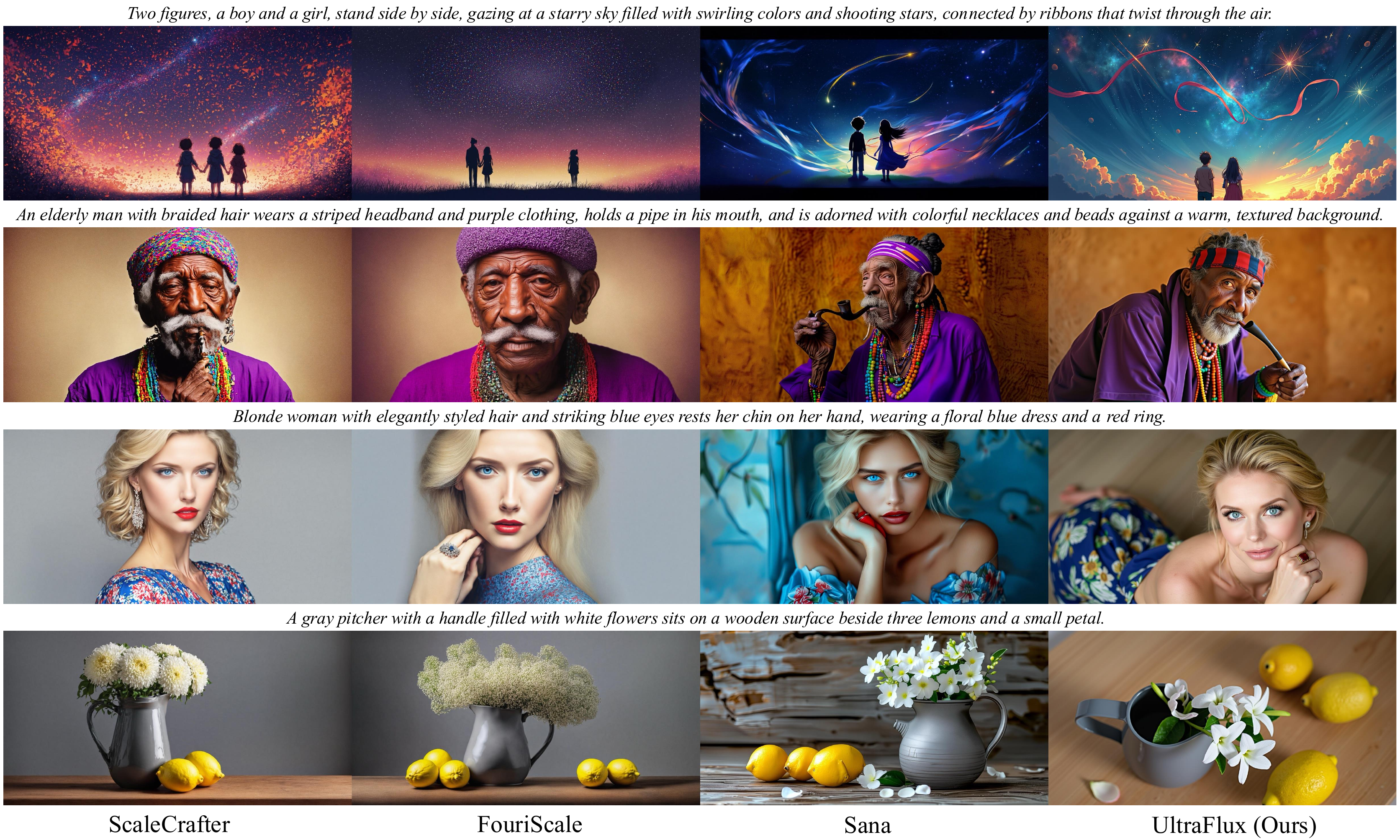}
\caption{Visual comparison of open-source methods at 1:2 aspect ratio (2048x4096).}
\label{fig:more_1-2_visual_comparison}
\end{figure*}

\begin{figure*}[t!]
\centering
\includegraphics[page=2, width=0.92\linewidth]{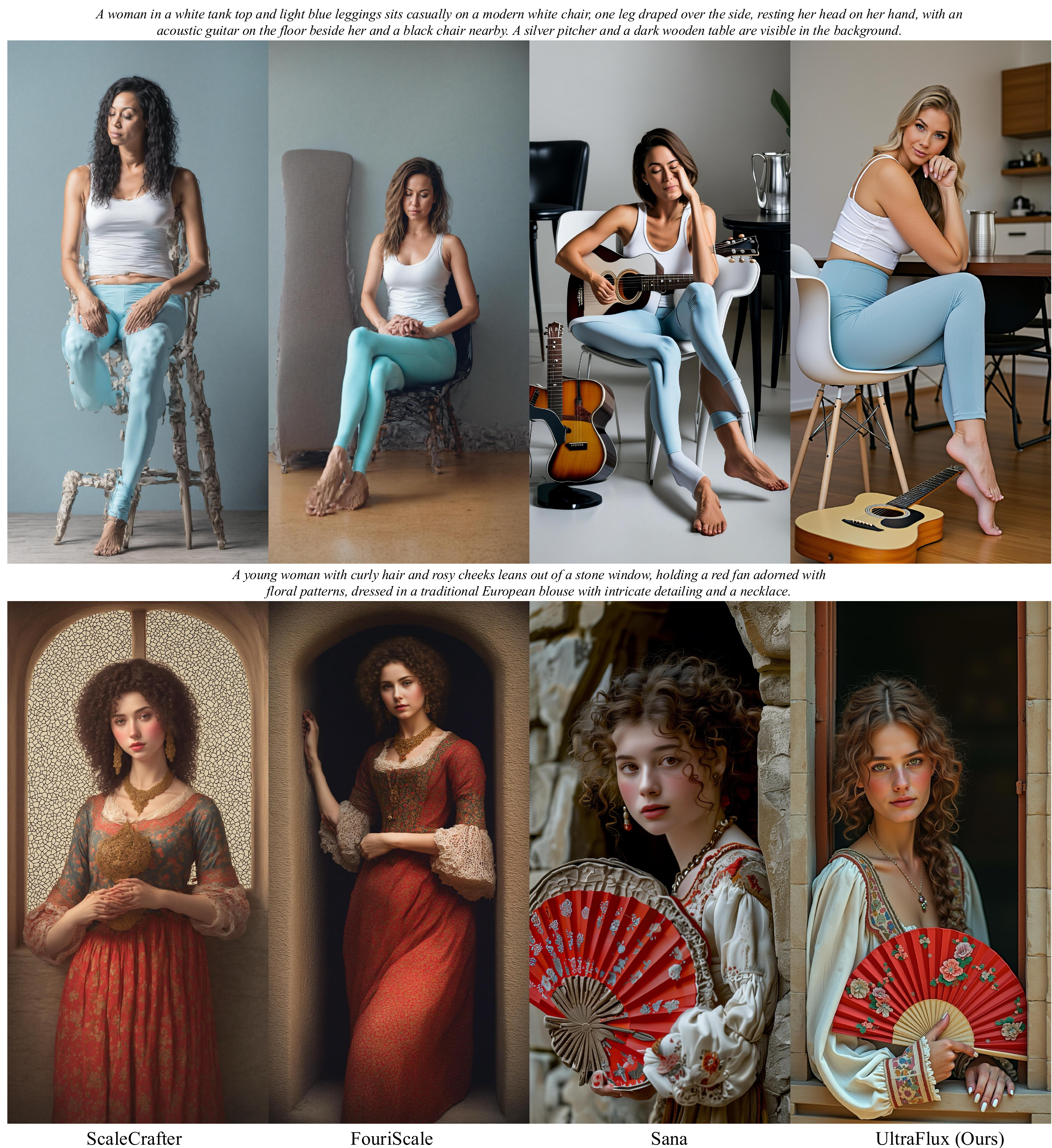}
\caption{Visual comparison of open-source methods at 2:1 aspect ratio (4096x2048).}
\label{fig:more_2-1_visual_comparison}
\end{figure*}

\begin{figure*}[t!]
\centering
\includegraphics[page=3, width=0.92\linewidth]{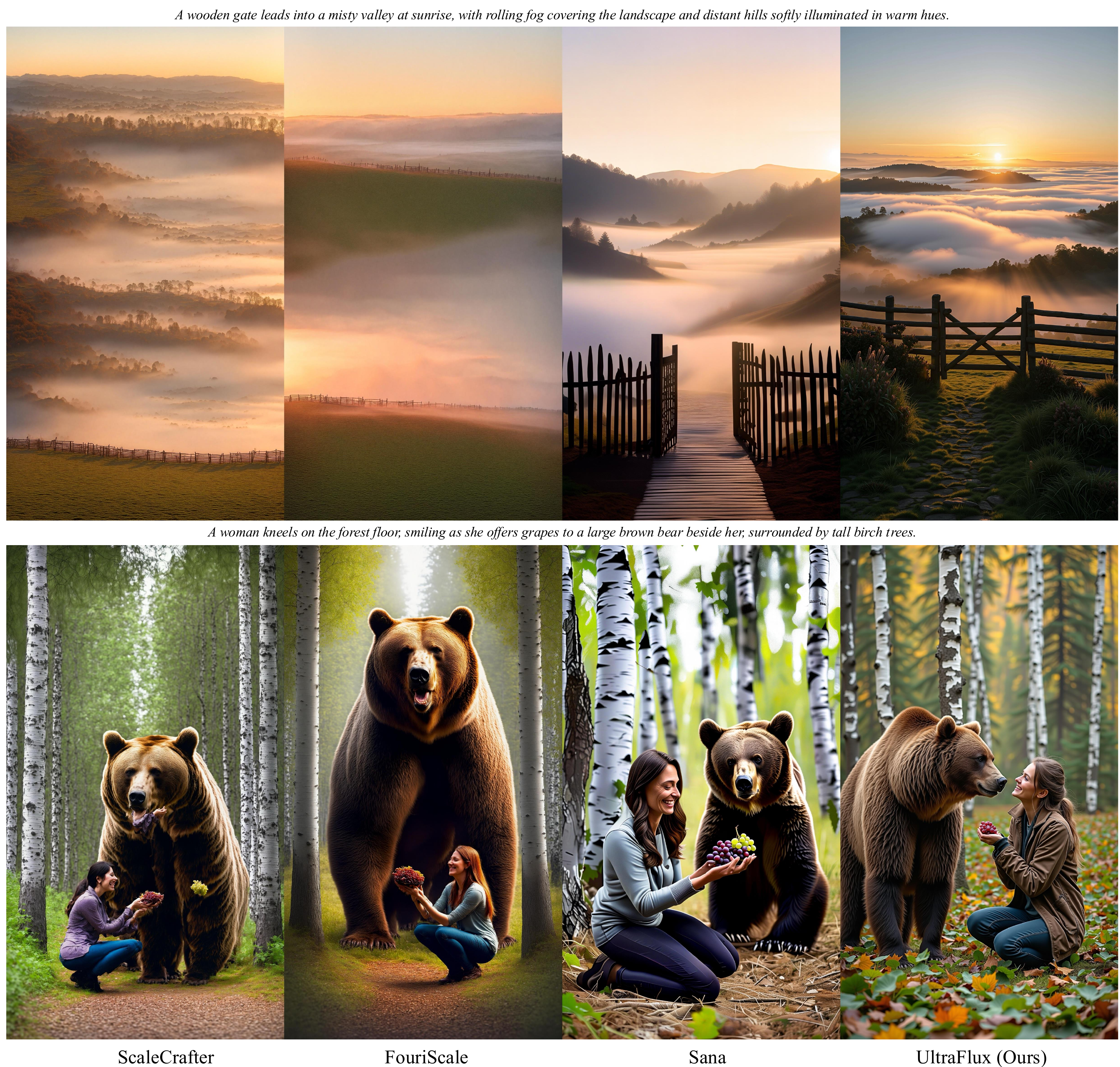}
\caption{Visual comparison of open-source methods at 16:9 aspect ratio (5120x2880).}
\label{fig:more_16-9_visual_comparison}
\end{figure*}

\begin{figure*}[t!]
\centering
\includegraphics[page=4, width=0.92\linewidth]{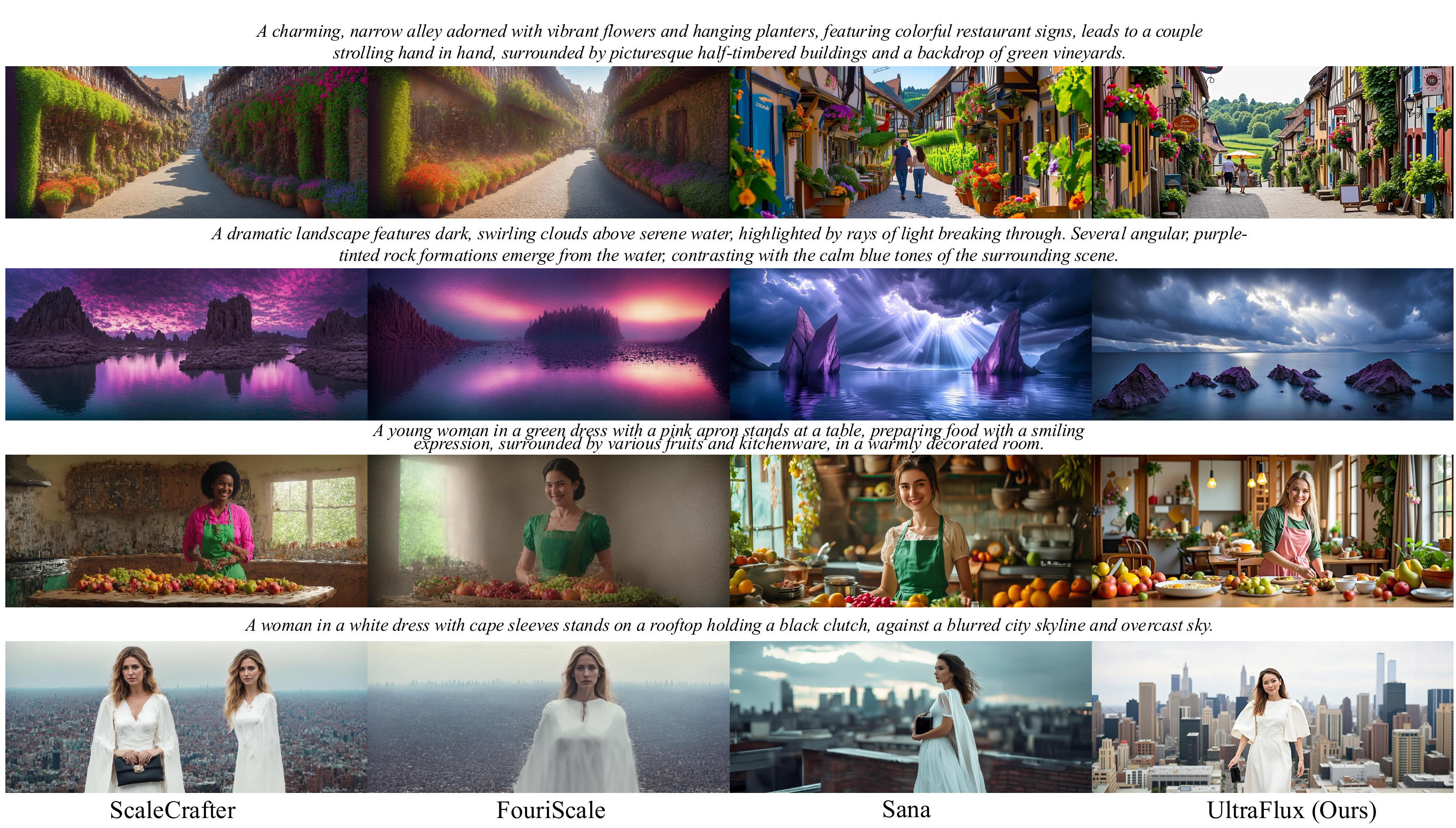}
\caption{Visual comparison of open-source methods at 1:2.39 aspect ratio (2496x5952).}
\label{fig:more_1-2.39_visual_comparison}
\end{figure*}

This section provides visual comparisons with SOTA methods at wide aspect ratios, complementing our quantitative analysis in Table~\ref{tab:more_multi_ar}. The results are presented in Fig.~\ref{fig:more_1-2_visual_comparison} (1:2), Fig.~\ref{fig:more_2-1_visual_comparison} (2:1), Fig.~\ref{fig:more_16-9_visual_comparison} (16:9) and Fig.~\ref{fig:more_1-2.39_visual_comparison} (1:2.39), respectively.

At the 1:2 aspect ratio, all methods produce visually plausible results without severe artifacts. However, our results are more visually appealing with better composition and aesthetic quality. In the 2:1 case, methods such as Scalecrafter and Fouriscale exhibit noticeable structural distortions and artifacts, while Sana also shows visible flaws. In contrast, our method generates remarkably natural and coherent images. At the 2.39:1 ultra-wide ratio, both Scalecrafter and Fouriscale suffer from mild misalignment with text prompts as well as detail degradation. Our results not only avoid these issues but also outperform Sana in overall visual quality.

These observations demonstrate that our approach consistently maintains state-of-the-art performance and high visual fidelity across a spectrum of challenging aspect ratios.

\section{Additional Visual Comparison With Open-Source methods}
\begin{figure*}[t!]
\centering
\includegraphics[page=1, width=0.92\linewidth]{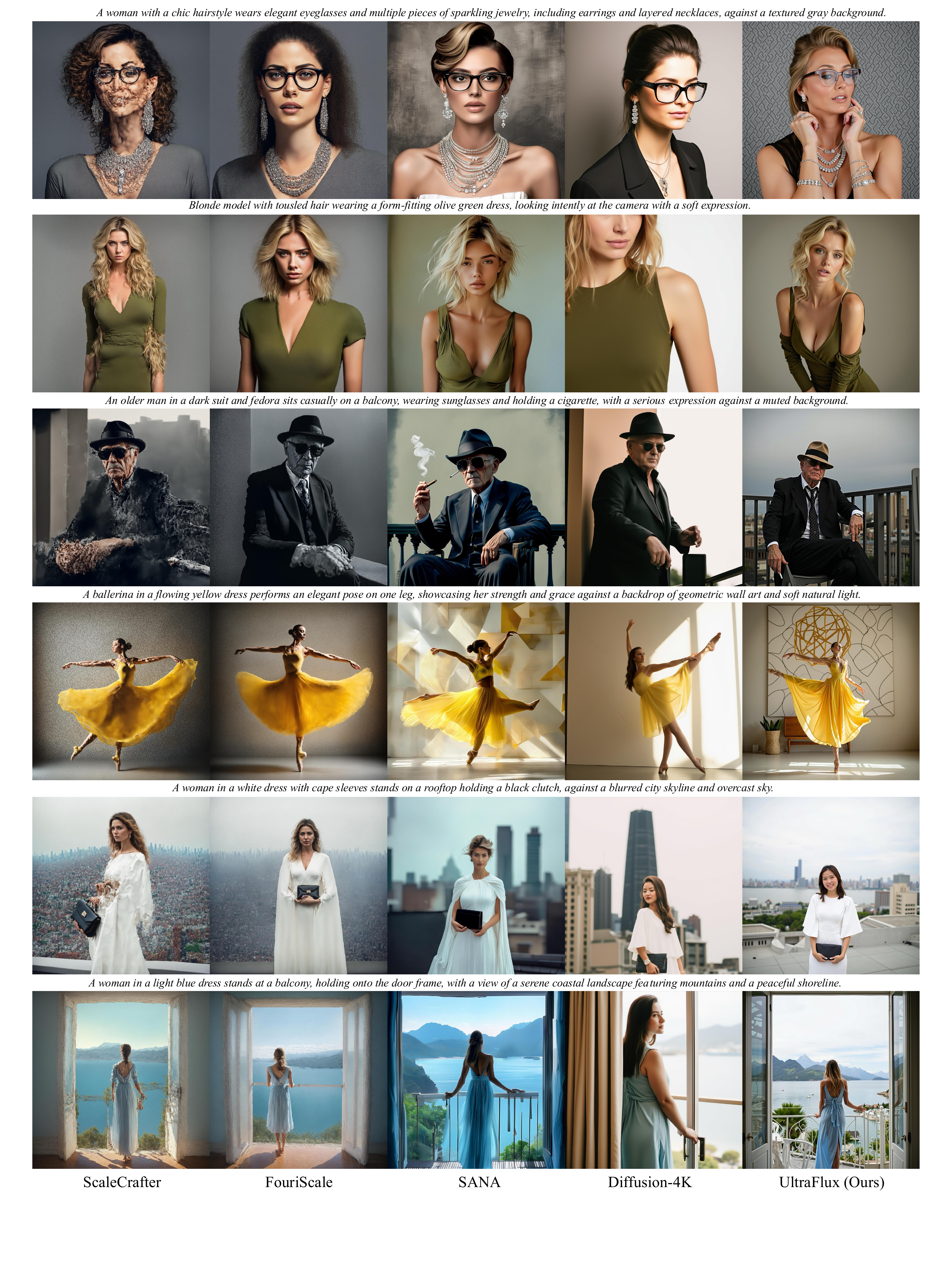}
\caption{More visual comparison of open-source methods on the Aesthetic-Eval@4096 benchmark at 4096$\times$4096 resolution.}
\label{fig:more_1-1_visual_comparison-1}
\end{figure*}

\begin{figure*}[t!]
\centering
\includegraphics[page=2, width=0.92\linewidth]{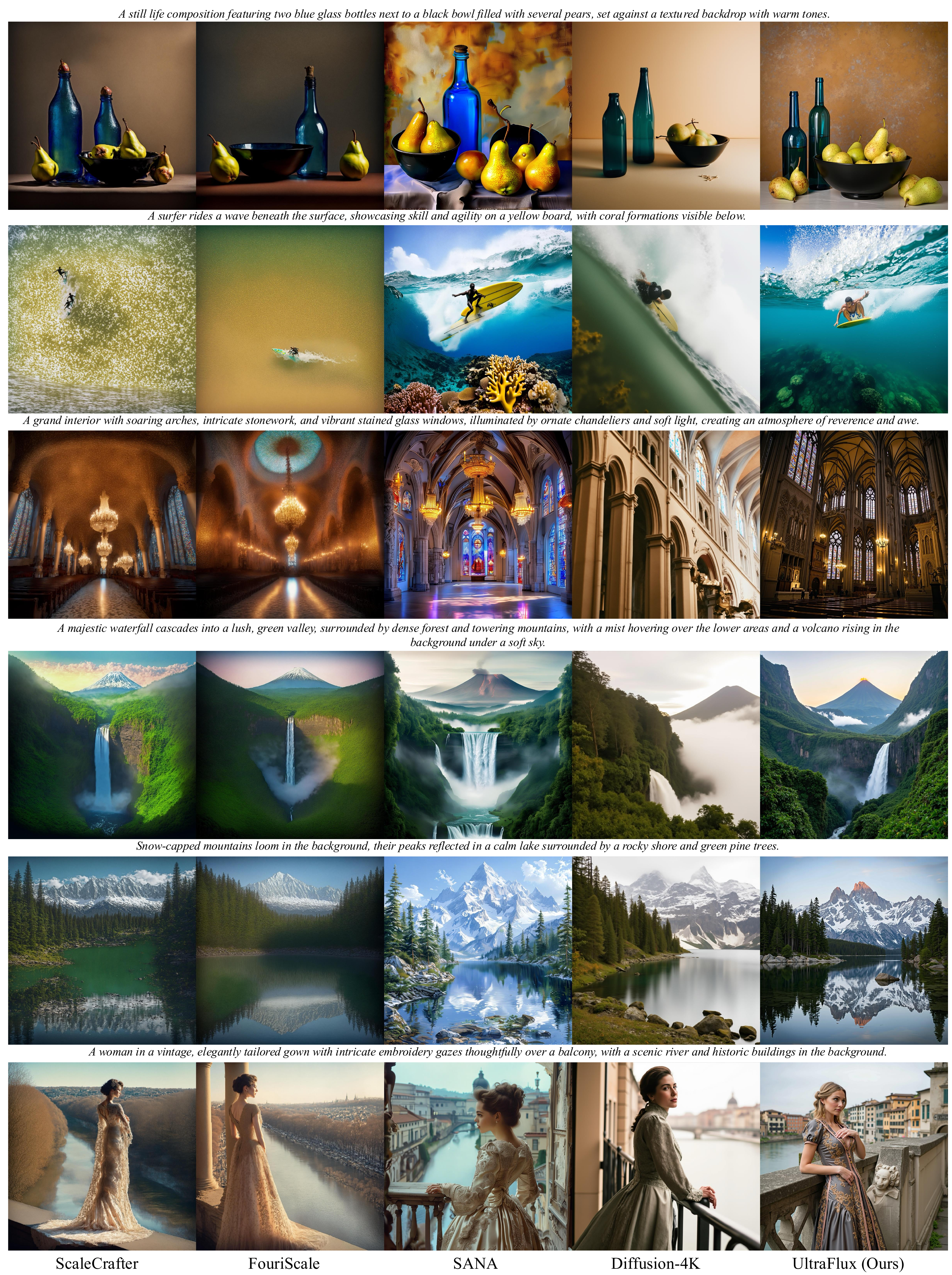}
\caption{More visual comparison of open-source methods on the Aesthetic-Eval@4096 benchmark at 4096$\times$4096 resolution.}
\label{fig:more_1-1_visual_comparison-2}
\end{figure*}

\begin{figure*}[t!]
\centering
\includegraphics[page=3, width=0.92\linewidth]{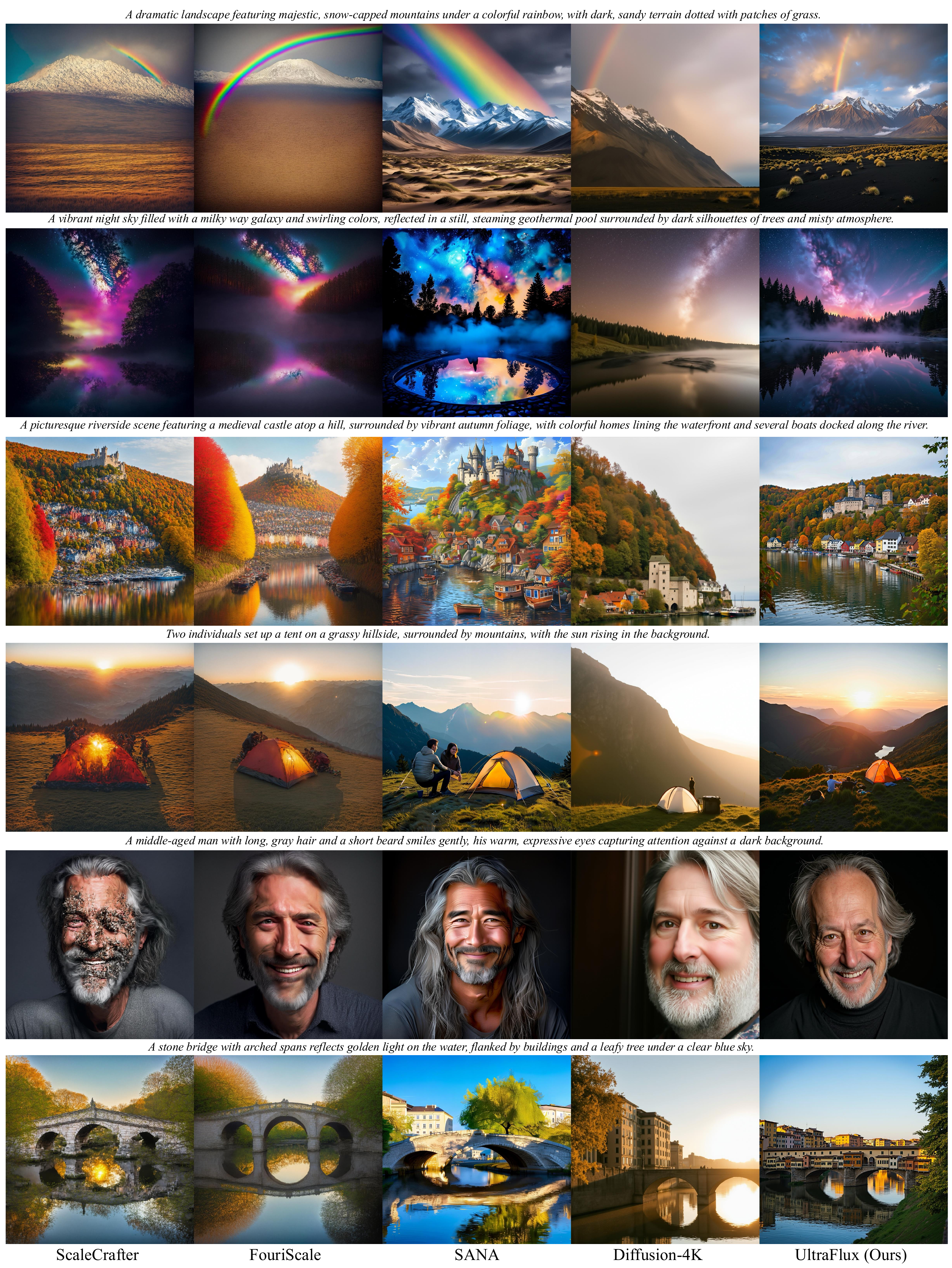}
\caption{More visual comparison of open-source methods on the Aesthetic-Eval@4096 benchmark at 4096$\times$4096 resolution.}
\label{fig:more_1-1_visual_comparison-3}
\end{figure*}

In Figures~\ref{fig:more_1-1_visual_comparison-1}--\ref{fig:more_1-1_visual_comparison-3}, additional visual comparisons are provided. From the results, we observe that ScaleCrafter sometimes produces images with noticeable distortions, while FouriScale occasionally struggles to fully capture textual content. The images generated by SANA, on the other hand, can appear somewhat overly smoothed or "oily." In contrast, compared to Diffusion-4K, our method consistently delivers higher-quality images with more visually appealing results, offering a more pleasant overall experience.

\section{Limitations}
Although UltraFlux substantially improves native-4K, multi-AR generation over prior open-source baselines, the system still has several practical limitations.

\noindent\textbf{Sampling cost and memory footprint.}
First, UltraFlux is not yet a \emph{efficient} 4K generator. Even with the F16 VAE and our optimized DiT backbone, sampling at native 4K with 50–60 flow-matching steps remains noticeably slower than 1K-class models and requires a high-end 50GB GPU to avoid aggressive offloading. This compute and memory footprint limits deployment to research- or data-center–grade hardware, and makes large-scale 4K sampling expensive compared to lower-resolution pipelines or distilled student models.

\noindent\textbf{Aesthetic ceiling and robustness.}
Second, while our data–model co-design delivers consistent gains in automatic metrics and Gemini-based preference studies, the aesthetic quality is not uniformly top-tier across all prompts and domains. In challenging cases, UltraFlux can still produce occasional over-smoothed textures, minor geometric artifacts, or compositions that are less polished than those from heavily engineered proprietary systems. Our co-design focuses on the 4K + multi-AR regime rather than absolute peak aesthetics, and there remains headroom for further preference alignment, prompt understanding, and content diversity.

\noindent\textbf{Scope of co-design.}
Finally, the present work primarily co-designs dataset, positional encoding, VAE, and loss under a single large DiT backbone. We do not address complementary axes such as sparse or low-rank attention, lightweight decoders, or distillation to smaller 4K models, which could significantly reduce memory usage and latency. Extending UltraFlux-style co-design to more parameter-efficient architectures and to broader data domains (e.g., specialized scientific or medical imagery) is an important direction for future work.

\section{Details About Gemini-based Preference Evaluation.}
In this section, we provide additional details on the Gemini-based preference evaluation used to assess the visual quality and prompt alignment of different models. As part of this evaluation, Gemini-2.5-Flash, in reasoning mode, is employed to judge image pairs based on their aesthetic appeal and alignment with the given prompt. The following is an example of the exact prompt used for evaluating \emph{aesthetic preferences} in our study. For each image pair, Gemini is asked to assess various aspects such as composition, sharpness, lighting, and overall visual appeal, ensuring that the evaluation process is both consistent and reproducible.

\begin{example}{Pairwise Preference for Aesthetics}
You are an impartial image aesthetics judge. Compare Image A and Image B, and decide which one better fits human aesthetic preferences overall. 

Evaluate:
\begin{itemize}[left=0pt,nosep]
    \item Composition
    \item Sharpness / clarity
    \item Lighting / contrast
    \item Color harmony
    \item Noise / compression artifacts
    \item Overall visual appeal
\end{itemize}

Be decisive; only return \texttt{"tie"} if the two images are nearly identical in quality.

\textbf{Return strictly in the following JSON format (no explanations, no extra text):}
\begin{lstlisting}[language=json]
{
    "preferred": "A | B | tie",
    "a_score": 0-100,
    "b_score": 0-100,
    "reasons": "short explanation"
}
\end{lstlisting}
\end{example}

\section{Details About Prompt Refiner using GPT-4O.}

\begin{example}{GPT-4O Prompt Refining Process}
\textbf{System prompt:}
You are a senior prompt refiner for AI image generation. Expand each short prompt into a single rich, high-aesthetic prompt.

Requirements:
\begin{itemize}[left=0pt,nosep]
    \item Length: 55–100 words; one line per item; no newlines, numbering, or quotes.
    \item Preserve the original subject and intent; do not invent brands, copyrighted IP, or named people.
    \item Composition: camera angle, shot size/framing, focal length or lens type, foreground/midground/background, environment context.
    \item Subject attributes: age range, gender expression where implied, appearance details (hair/eyes/skin or material), clothing/fabric, pose, expression/action.
    \item Lighting and color: key light quality/direction, color temperature, time of day/season/weather, palette or dominant hues.
    \item Style/medium: photographic or cinematic unless the input implies another medium; mention film look or post-processing if appropriate.
    \item Quality: tasteful, coherent, non-repetitive language; avoid keyword stuffing.
\end{itemize}

\textbf{User prompt:}
Short prompts: \{list of input prompts\}

Language: Write outputs in \textit{[language]} (Chinese/English); one line per item. For each input, produce exactly one refined prompt; avoid lists, bullets, or line breaks inside items.

\textbf{Expected output format:}
\begin{lstlisting}[language=json]
[
  "Refined prompt 1",
  "Refined prompt 2",
  ...
]
\end{lstlisting}
\end{example}

To further refine the quality of input prompts, we employ GPT-4o as a front-end for our \emph{UltraFlux w. Prompt Refiner (Ours)} configuration. The process of prompt refinement involves transforming short and concise prompts into more detailed, high-aesthetic descriptions suitable for image generation tasks. The GPT-4o model expands each input prompt into a rich description, incorporating essential elements such as composition, lighting, subject attributes, and stylistic choices. The refined prompts follow a strict set of guidelines to maintain coherence, clarity, and aesthetic quality, ensuring that they meet the requirements for high-fidelity image generation. In the following example, we provide the exact system prompt and user instructions used to guide GPT-4o in refining a list of short prompts. The prompts are designed to ensure that the model generates visually appealing and contextually appropriate descriptions for each input. This prompt refining process ensures that the generated prompts are detailed, high-quality, and aligned with the intended visual aesthetics. The use of GPT-4o to refine short prompt significantly enhances the input quality, making it suitable for use in high-fidelity image generation tasks.

\clearpage
\end{document}